\documentclass[conference]{IEEEtran}
\pdfoutput=1
\IEEEoverridecommandlockouts
\usepackage{cite}
\usepackage{amsmath,amssymb,amsfonts}
\usepackage{pifont}
\usepackage{textcomp}
\usepackage[normalem]{ulem}
\usepackage{xcolor}
\usepackage{mathtools}
\usepackage{xr}
\usepackage[linesnumbered,ruled,vlined]{algorithm2e}
\usepackage{empheq}
\usepackage{cases} 
\usepackage{microtype}
\usepackage{subcaption}
\usepackage[font={small}]{caption}
\usepackage{balance}
\usepackage{bibunits}
\usepackage{booktabs}
\usepackage[hidelinks]{hyperref}
\usepackage{multirow}
\usepackage{float}
\usepackage{bibunits}
\usepackage[switch]{lineno}
\def\BibTeX{{\rm B\kern-.05em{\sc i\kern-.025em b}\kern-.08em
    T\kern-.1667em\lower.7ex\hbox{E}\kern-.125emX}}

\DeclareMathOperator*{\argmin}{arg\,min}

\usepackage{array}
\usepackage{etoolbox}
\newcolumntype{P}[1]{>{\centering\arraybackslash}p{#1}}%
\SetKwInput{KwInput}{Input}                %
\SetKwInput{KwOutput}{Output}              %
\SetKw{Break}{break}

\newcommand*{\defeq}{\stackrel{\text{def}}{=}}
\SetCommentSty{mycommfont}

\SetAlgoSkip{}
\makeatletter
\def\footnoterule{\kern-3\p@
  \hrule \@width 2in \kern 2.6\p@} %
\makeatother

\usepackage{color}

\begin{document}

\pdfinfo{
/Robust and Explainable Autoencoders for Time Series Outlier Detection---Extended Version
/Tung Kieu, Bin Yang, Chenjuan Guo, Christian S. Jensen, Yan Zhao, Feiteng Huang, Kai Zheng}

\title{Robust and Explainable Autoencoders for Unsupervised Time Series Outlier Detection---Extended Version}

\author{
	\IEEEauthorblockN{Tung Kieu$^{1}$, 
	Bin Yang$^1$, 
	Chenjuan Guo$^1$, 
	Christian S. Jensen$^1$, 
	Yan Zhao$^1$, 
	Feiteng Huang$^2$, 
	Kai Zheng$^3$
	}
	\IEEEauthorblockA{$^1$Aalborg University, Denmark~~ 
		$^2$Huawei Cloud Database Innovation Lab, China~~\\
		$^3$University of Electronic Science and Technology of China, China\\
		\{tungkvt, byang, cguo, csj, yanz\}@cs.aau.dk, huangfeiteng@huawei.com, zhengkai@uestc.edu.cn
	}
}

\newcommand{\edit}[1]{{\textcolor{red}{#1}}}
\newcommand{\rev}[1]{{\textcolor{black}{#1}}}

\maketitle
\thispagestyle{plain}
\pagestyle{plain}

\begin{abstract}
\rev{Time series data occurs widely, and outlier detection is a fundamental problem in data mining, which has numerous applications.} Existing autoencoder-based approaches deliver state-of-the-art performance on challenging real-world data but are vulnerable to outliers and exhibit low explainability. To address these two limitations, we propose robust and explainable unsupervised autoencoder frameworks that decompose an input time series into a clean time series and an outlier time series using autoencoders. Improved explainability is achieved because clean time series are better explained with easy-to-understand patterns such as trends and periodicities. We provide insight into this by means of a post-hoc explainability analysis and empirical studies. In addition, since outliers are separated from clean time series iteratively, our approach offers improved robustness to outliers, which in turn improves accuracy. We evaluate our approach on five real-world datasets and report improvements over the state-of-the-art approaches in terms of robustness and explainability.

This is an extended version of ``Robust and Explainable Autoencoders for Unsupervised Time Series Outlier Detection"~\cite{tungicde2022second}, to appear in IEEE ICDE 2022.
\end{abstract}
\section{Introduction}
Analyses of time series yield knowledge of the underlying processes that generate the time series and in turn enable us to understand those processes~\cite{razvanicde2021,Razvanicde2022,DBLP:conf/icde/Hu0GJX20,wupvldb,MileTS}. 
Time series outlier detection is a fundamental problem in data mining and has been applied in many applications, including in transportation, crowdsourcing, and manufacturing~\cite{DBLP:journals/pvldb/PedersenYJ20,
tkdesean, DBLP:conf/waim/YuanSWYZY10, 
DBLP:journals/vldb/GuoYHJC20,DBLP:journals/vldb/PedersenYJ20,DBLP:conf/icde/LiuJYZ18}. 
Recent neural network based methods for outlier detection in time series achieve strong performance on challenging datasets.  %
These methods learn long-term, nonlinear temporal relationships in the data, outperforming existing, non-deep methods based on similarity search~\cite{DBLP:conf/icdm/YehZUBDDSMK16} and density-based clustering~\cite{DBLP:conf/sigmod/BreunigKNS00}. %
Neural network based methods  
first compress an input time series %
into a {compact} latent representation using an encoder and then reconstruct the original time series from the latent representation using a decoder. 
This encoder-decoder architecture, called an autoencoder~\cite{DBLP:jour/science/HinSal06,DBLP:conf/cikm/Kieu0GJ18}, imposes an information bottleneck~\cite{tishby2000information} that encourages the compact latent representation to capture only the most representative patterns of the input time series while disregarding {non-representative patterns such as outliers}. 
This makes it possible to detect outliers by measuring the reconstruction error between the input time series and the reconstructed time series. The larger the reconstruction errors are, the more likely it is that the corresponding observations in the time series are outliers. 
This makes autoencoders fit unsupervised outlier detection, as they do not rely on any outlier labels that indicate which observations are outliers, but rely purely on reconstruction errors. Although autoencoders achieve competitive performance, they face two challenges.

\begin{figure*}[ht]
	\begin{subfigure}[t]{.325\linewidth}
		\centering
		\includegraphics[width=1.0\linewidth]{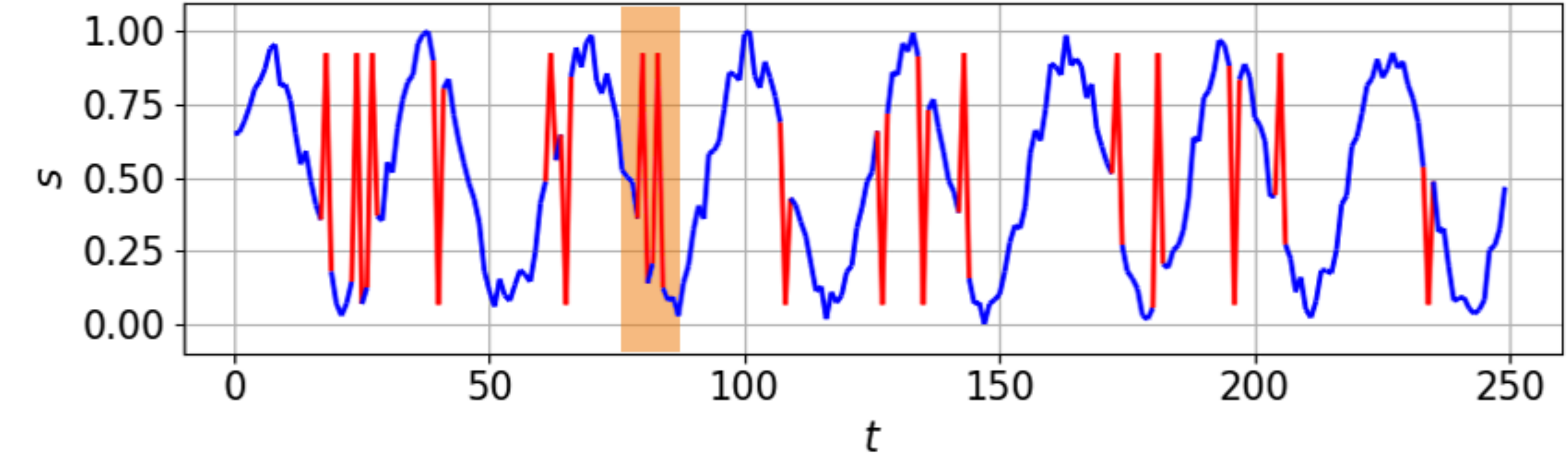}
		\caption{Input time series $\mathcal{T}$. Red spikes are outliers.}
		\label{fig:denosing_outlier}
	\end{subfigure}
	\label{fig:synthetic}
	\centering
	\begin{subfigure}[t]{.325\linewidth}
		\centering
		\includegraphics[width=1.0\linewidth]{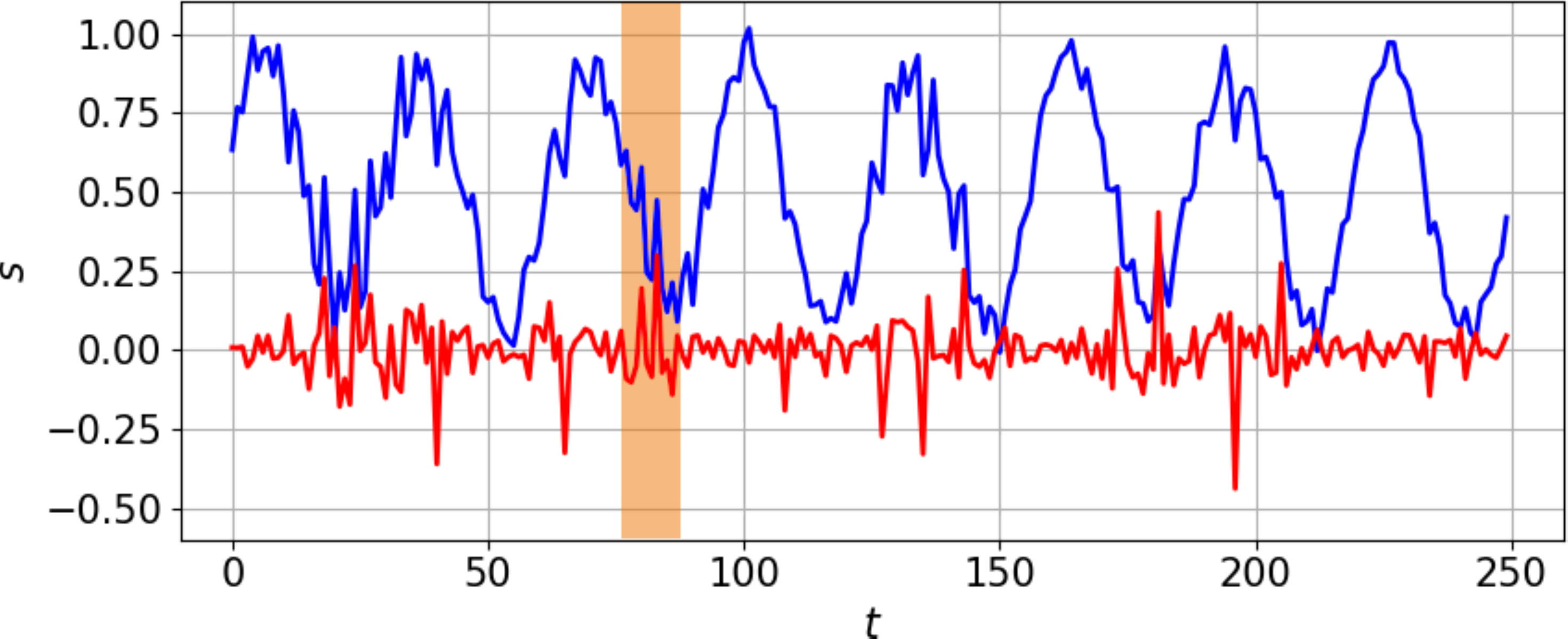}
		\caption{Standard autoencoder, \texttt{RNNAE}.}
		\label{fig:denosing_rnnae}
	\end{subfigure}
	\begin{subfigure}[t]{.325\linewidth}
		\centering
		\includegraphics[width=1.0\linewidth]{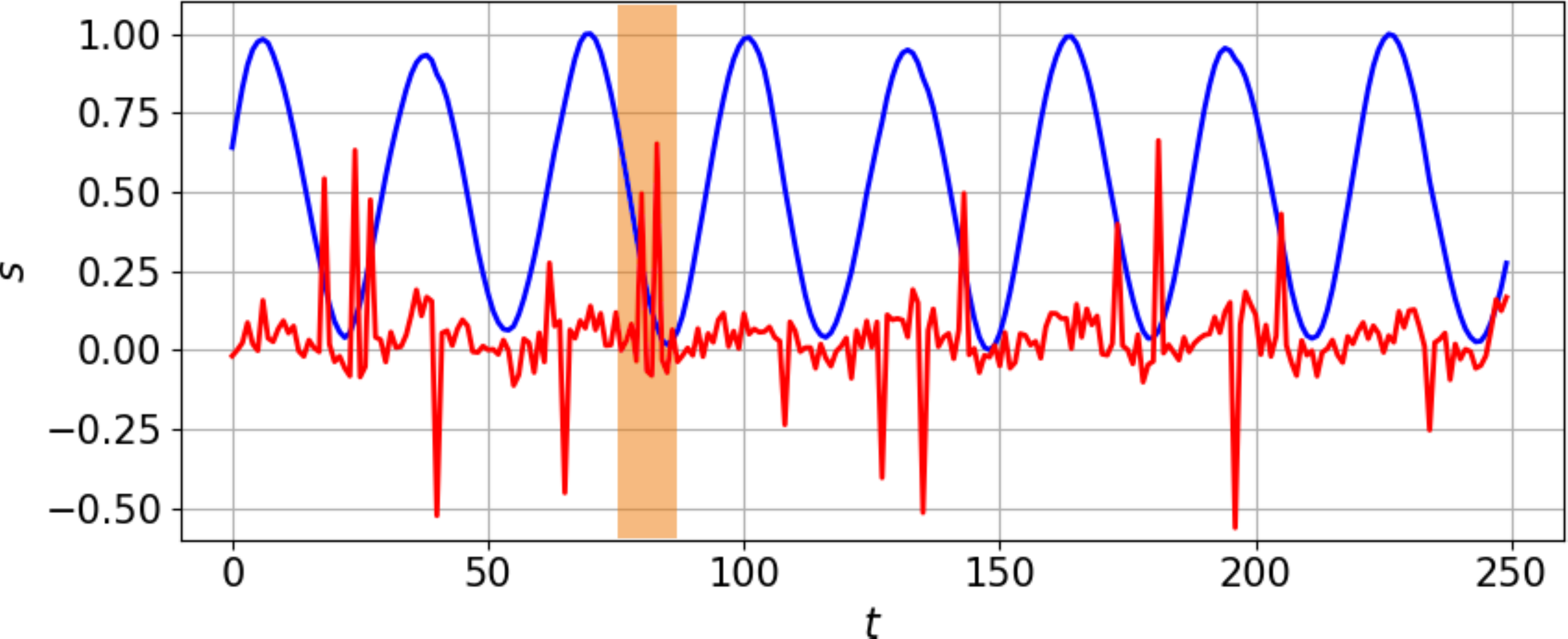}
		\caption{Robust autoencoder, \texttt{RDAE}.}
		\label{fig:denoising_rdae}
	\end{subfigure}
	\caption{Robustness and Explainability Illustration. In (b) and (c), the blue curves indicate the reconstructed time series $\mathcal{\hat{T}}$ returned by \texttt{RNNAE} and \texttt{RDAE}, and the red curves indicate the %
	reconstruction errors 
	$\mathcal{T}-\mathcal{\hat{T}}$. 
	\emph{Robustness: }outliers generally have larger reconstruction errors in (c) than in (b), e.g., in the highlighted orange region, suggesting that \texttt{RDAE} is more robust to outliers. 
	\emph{Explainability: }the blue curve in (c) has a more clear periodic pattern and much less hard-to-explain variations than in (b), suggesting that \texttt{RDAE} has higher explainability than \texttt{RNNAE}. %
	}
	\label{fig:robustness_example}
\end{figure*}

\noindent
(i) {\bf Robustness:} %
In unsupervised settings, the training data may already include outliers. 
Since encoders compress all observations
in an input time series, including outliers, %
the resulting latent
representations are sensitive to outliers. \rev{
Small numbers of outliers may still contaminate the latent representations, especially if their magnitudes are large.}
In other words, there is a risk that outliers in the training data pollute the latent representations such that the latent representations also capture the outlier patterns; thus, some outliers may have small reconstruction errors, which are then difficult to separate from clean data. 
This adversely affects accuracy. For example, %
the blue curve %
in 
Fig.~\ref{fig:denosing_rnnae} shows a reconstructed time series from a polluted latent representation.  This yields small reconstruction errors for some outliers, making them difficult to detect (see the orange region in Fig.~\ref{fig:denosing_rnnae}). %
To avoid this, robust solutions are called for such that the latent representations are less affected by the outliers in the training data. %

\noindent
(ii) {\bf Explainability:} 
Autoencoders regard observations with large reconstruction errors as outliers. Put differently, given an input time series $\mathcal{T}$, autoencoders regard the reconstructed time series $\mathcal{\hat{T}}$ as the ``clean'' time series that should occur if the underlying system generates the time series in a normal state. %
If observations from the input time series deviate substantially from the corresponding observations in the reconstructed time series, meaning that the corresponding reconstruction errors $\mathcal{T}-\mathcal{\hat{T}}$ are large, then the autoencoders regard these observation as outliers. 
Thus, to understand why particular observations are regarded as outliers by autoencoders, it is important to understand
the corresponding reconstruction errors $\mathcal{T}-\mathcal{\hat{T}}$. Since the input time series $\mathcal{T}$ is given, it is then important to understand 
the key features of the reconstructed time series $\mathcal{\hat{T}}$, such as increasing or decreasing trends or different periodicities, as this offers information on which observations should occur in
a normal state, which %
helps users understand the reconstruction errors $\mathcal{T}-\mathcal{\hat{T}}$ that indicate outliers.
However, the reconstructed time series $\mathcal{\hat{T}}$ produced by existing autoencoders (e.g., the blue curve in Fig.~\ref{fig:denosing_rnnae}) are often complex with hard-to-explain specifics, making it difficult for users to understand which observations should occur in a normal state. 
This calls for more explainable solutions such that the reconstructed time series have %
a clear pattern and few hard-to-explain variations. %

We address these two limitations by providing two novel autoencoder frameworks that improve robustness and explainability. %
Instead of reconstructing an input time series $\mathcal{T}$ directly, we decompose it into two: a clean time series $\mathcal{T}_\mathbf{L}$ and an outlier time series $\mathcal{T}_\mathbf{S}$. 
We then employ autoencoders to reconstruct only the clean time series $\mathcal{T}_\mathbf{L}$, thus preventing outliers from polluting the latent representation. This improves robustness, which in turn improves accuracy. 
Fig.~\ref{fig:denoising_rdae} shows the results of the proposed robust dual autoencoder \texttt{RDAE}, where the outliers in the orange region have larger reconstruction errors and are thus easier to detect. 

Next, since the latent representation only captures clean time series, we expect our approach better explain the reconstructed time series %
in terms of trends and periodicities. 
For example, the blue curve in Fig.~\ref{fig:denoising_rdae} shows the clean, reconstructed time series obtained from the proposed robust dual autoencoder \texttt{RDAE}. It includes clear periodic patterns without random variations, making it much easier for users to understand which observations are expected to occur in a normal state and hence understanding the reconstruction errors that indicate outliers, 
compared to that of Fig.~\ref{fig:denosing_rnnae}. %

Finally, we propose a post-hoc explainability analysis method to quantify the explainability of autoencoder based outlier detection methods, which %
in turn facilitates quantitative comparisons of explainability among different autoencoder based outlier detection methods. 
Unlike in accuracy evaluation, where well-established metrics such as precision-recall exist, it lacks established metrics when evaluating the explainability of different autoencoder based outlier detection methods in a quantitative manner. 
Most studies evaluate explainability through visualizations (e.g., by contrasting Figs.~\ref{fig:denosing_rnnae} and~\ref{fig:denoising_rdae})---though intuitive, visualization based explainability evaluation fails to provide quantitative results. %
Motivated by explainable machine learning~\cite{DBLP:journals/cacm/DuLH20}, we %
adopt post-hoc explainability analyses to design %
two metrics that enable quantitative evaluation of autoencoder based outlier detection methods. 

To the best of our knowledge, this is the first study that focuses on improving both robustness and explainability of autoencoders for time series outlier detection. 
\rev{More specifically, we make three contributions. 
}

\begin{itemize}
	\item We propose two autoencoder frameworks for unsupervised time series outlier detection  that offer improved robustness and explainability.  
	\item We propose a post-hoc explainability analysis technique that enables quantification of the explainability of autoencoder based outlier detection methods. 
	\item We report on extensive empirical studies that offer insight into pertinent design properties of the two frameworks and that compare with baselines and the state-of-the-art approaches in terms of accuracy and explainability. 
\end{itemize}

The rest of the paper is organized as follows. 
Section~\ref{sec:preliminaries} covers preliminaries. %
Section~\ref{sec:methodology} proposes the two frameworks.
Section~\ref{sec:explainability} details the post-hoc explainability analysis. 
Section~\ref{sec:experiments} reports experimental results, Section~\ref{sec:related works} discusses related work, and Section~\ref{sec:conclusion} concludes the paper.

A preliminary version~\cite{tungicde2022second} has been accepted by the 38th IEEE International Conference on Data Engineering (ICDE 2022). This version offers more detailed technical specifics and additional experimental results.

\section{Preliminaries}
\label{sec:preliminaries}
\subsection{Time Series Outlier Detection} 
A time series $\mathcal{T} = \langle \mathbf{s}_1, \dots, \mathbf{s}_C \rangle$ is a sequence of $C$ observations, where each observation $\mathbf{s}_i\in\mathbb{R}^{D}$. If $D=1$, $\mathcal{T}$ is \textit{univariate}. If $D>1$, $\mathcal{T}$ is \textit{multivariate} (or \textit{multidimensional}).

Given a time series $\mathcal{T} = \langle \mathbf{s}_1, \mathbf{s}_2, \dots, \mathbf{s}_C \rangle$, we aim at computing an outlier score $\mathcal{OS}(\mathbf{s}_i)$ for each observation $\mathbf{s}_i$ such that the higher $\mathcal{OS}(\mathbf{s}_i)$ is, the more likely it is that observation $\mathbf{s}_i$ is an outlier. 
\rev{We make no assumptions whether outliers are point or collective outliers. If the outlier scores of continuous observations are high, these observations can be detected as a collective outlier.}

\subsection{Robust Principal Component Analysis} 
Given a matrix $\mathbf{M}$, Principal Component Analysis (\texttt{PCA}) is able to identify a low rank matrix to approximate matrix $\mathbf{M}$. 
Since \texttt{PCA} often employs Singular Value Decomposition~(\texttt{SVD}) to identify the low-rank matrix, \texttt{PCA} also has the same problem as \texttt{SVD} of being very sensitive to outliers. 
To improve the performance of \texttt{PCA} when outliers exist, Robust Principal Component Analysis (\texttt{RPCA})~\cite{DBLP:journals/jacm/CandesLMW11} has been proposed.   %
\texttt{RPCA} aims to separate the given matrix $\mathbf{M}$ into the sum of two matrices---a low-rank matrix $\mathbf{L}$ that represents the clean data and a matrix $\mathbf{S}$ that consists of element-wise outliers. 
Specifically, \texttt{RPCA} decomposes the original matrix $\mathbf{M}$ such that $\mathbf{M} = \mathbf{L} + \mathbf{S}$. Here, $\mathbf{L}$ is a low-rank matrix that aims to approximate the clean data in the original matrix $\mathbf{X}$, and $\mathbf{S}$ is a sparse matrix that consists of element-wise outliers that should not be captured by the low-rank matrix $\mathbf{L}$. 
In other words, if the low rank matrix $\mathbf{L}$ tries to capture the outliers in $\mathbf{S}$ then $\mathbf{L}$ is skewed to the outliers and thus cannot capture appropriately the clean data. 
\texttt{RPCA} achieves the decomposition by solving the optimization problem shown in Eq.~\ref{fn: optimization_RPCA}.

\begin{equation}
	\begin{aligned}
		\argmin_{\mathbf{L}, \mathbf{S}} \; & \mathrm{rank}(\mathbf{L}) + \lambda||\mathbf{S}||_0 \; \text{s.t.} \; \mathbf{X} = \mathbf{L} + \mathbf{S}
	\end{aligned}
	\label{fn: optimization_RPCA}
\end{equation}

Here, $\mathrm{rank}(\mathbf{L})$ is the rank of matrix $\mathbf{L}$; $||\mathbf{S}||_0$ is the $\ell_0$ norm of matrix $\mathbf{S}$, which counts the number of non-zero elements in $\mathbf{S}$; and $\lambda$ is a coefficient that controls the relative importance of $||\mathbf{S}||_0$. In addition, the optimization is constrained by $\mathbf{M} = \mathbf{L} + \mathbf{S}$ because $\mathbf{M}$ is decomposed into $\mathbf{L}$ and $\mathbf{S}$. 
Minimizing the loss function makes it possible to identify a low rank matrix $\mathbf{L}$ that approximates the original matrix $\mathbf{M}$ and a sparse matrix $\mathbf{S}$ that includes outliers. 

Although \texttt{RPCA} is able to effectively identify and remove outliers, it does not support time series directly. An additional limitation %
is that only linear transformations are employed. 

\subsection{Autoencoders}

An autoencoder (\texttt{AE})~\cite{DBLP:jour/science/HinSal06} consists of an encoder $E_{\theta_{AE}}(\cdot)$ and a decoder $D_{\theta_{AE}}(\cdot)$. The encoder $E_{\theta_{AE}}$ takes as input $\mathbf{M} \in \mathbb{R}^{m}$ and maps it into a compressed representation $\mathbf{H} \in \mathbb{R}^{n}$, where $n \ll m$. This occurs at the so-called \emph{bottleneck layer}.
Then, the decoder $D_{\theta_{AE}}$ takes $\mathbf{H}$ as input and outputs $\hat{\mathbf{M}}\in \mathbb{R}^{m}$, such that $\hat{\mathbf{M}}$ is as similar as possible to $\mathbf{M}$. This process is expressed in Eq.~\ref{eq:obj}, and Fig.~\ref{fig:AE} offers a framework overview of \texttt{AE}.

\begin{align}
	\argmin_{\theta_{AE}} ||\mathbf{M} - \hat{\mathbf{M}} ||_2 = \argmin_{\theta_{AE}} ||\mathbf{M} - D_{\theta_{AE}}(E_{\theta_{AE}}(\mathbf{M}))||_2,
	\label{eq:obj}
	\vspace{-0.5em}
\end{align}

where $\theta_{AE}$ are the learnable parameters of the \texttt{AE}. %
When using neural networks, non-linear transformations can be added easily into the encoder and decoder, thus offering opportunities to solve the limitation in \texttt{RPCA} of only capturing linear relationships. In addition, there exist neural networks that capture well time series.

\subsection{Design Considerations}

We summarize the key design considerations underlying the different approaches 
in Table~\ref{tab:advantages}. 
First, we consider whether a method supports time series data, noting that \texttt{RPCA} does not directly support time series data. 
Second, we consider whether a method is robust to outliers, which \texttt{AE}s are not because they work on all the data and do not distinguish outliers from clean data. 
Third, we consider explainability. \texttt{AE}s exhibit low explainability, while \texttt{RPCA} has strong theoretical underpinnings. 
Fourth, we consider whether a method is capable of supporting non-linear relationships, which often appear in complex time series. Here, \texttt{RPCA} falls short because it only uses linear operations and thus is only able to capture linear relationships. 
\rev{Fifth, we consider multi-view learning mechanism, which is not supported by any of \texttt{RPCA} or \texttt{AE}s.}
Based on the above analysis, we proceed to propose two robust and explainable autoencoder frameworks that achieve all the design considerations. 

\begin{table}[!htp]
	\centering
	\caption{Summary of key design considerations.}
	\label{tab:advantages}
	\begin{tabular}{|p{0.6cm}|P{1.3cm}P{0.6cm}P{1.2cm}P{1.3cm}P{1.3cm}|}%
		\toprule
		& \textbf{Time series} & \textbf{Robust} & \textbf{Explainable} & \textbf{Non-linear} & \textbf{\rev{Multi-view}}\\
		\midrule
		\texttt{RPCA} & \ding{55} & \ding{51} & {High} & \ding{55} & \rev{\ding{55}} \\
		\texttt{AE} &  \ding{51} & \ding{55} & {Low} & \ding{51} & \rev{\ding{55}} \\ \hline
		\textbf{Ours} &  \ding{51} & \ding{51} & {High} & \ding{51} & \rev{\ding{51}} \\
		\bottomrule
	\end{tabular}
\end{table}

\section{Methodology}
\label{sec:methodology}
\subsection{Overall Idea}
Although \texttt{RPCA} effectively identifies outliers, no attempt at using \texttt{RPCA}  for time series outlier detection exist. Following the principles of \texttt{RPCA}, we propose neural net based \texttt{AE}s to decompose time series $\mathcal{T}$ into a clean time series $\mathcal{T}_\mathbf{L}$ and an outlier time series $\mathcal{T}_\mathbf{S}$ such that $\mathcal{T} = \mathcal{T}_\mathbf{L} + \mathcal{T}_\mathbf{S}$.

We expect $\mathcal{T}_\mathbf{L}$ to represent typical patterns of the underlying system that generates the time series, e.g., trends and periodicities. Then, the outlier time series $\mathcal{T}_\mathbf{S}$ is expected to include the data points that cannot be captured by the typical patterns. This offers explainability of the outlier detection process. 

Since patterns in time series are often nonlinear, e.g., having different periodicities, we employ deep \texttt{AE}s with nonlinear activation functions. Specifically, we propose two \texttt{AE} frameworks that differ in how a time series is decomposed. The first framework employs a single \texttt{AE} to decompose the input time series $\mathcal{T}$ into clean and outlier parts. 
The second framework employs two \texttt{AE}s to decompose $\mathcal{T}$ from two different views. It first converts $\mathcal{T}$ into a lagged matrix $\mathbf{M}$. 
Then, the two \texttt{AE}s make the decomposition from the matrix vs. time series views, respectively. %
We proceed to elaborate the two frameworks. 

\begin{figure*}[t]
\centering
\begin{minipage}{.25\textwidth}
  \centering
  \includegraphics[width=.9\linewidth]{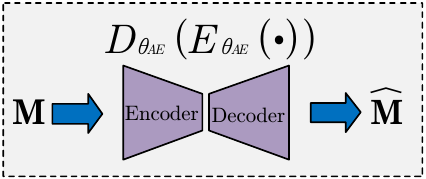}
  \captionof{figure}{\texttt{AE} framework.}
  \label{fig:AE}
\end{minipage}%
\begin{minipage}{.25\textwidth}
  \centering
  \includegraphics[width=.9\linewidth]{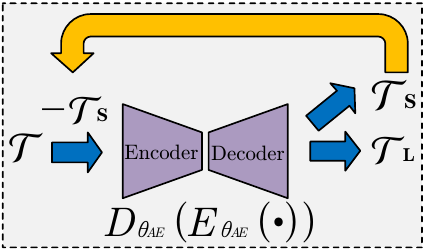}
  \captionof{figure}{\texttt{RAE} framework.}
  \label{fig:RAE}
\end{minipage}%
\begin{minipage}{.5\textwidth}
  \centering
  \includegraphics[width=1\linewidth]{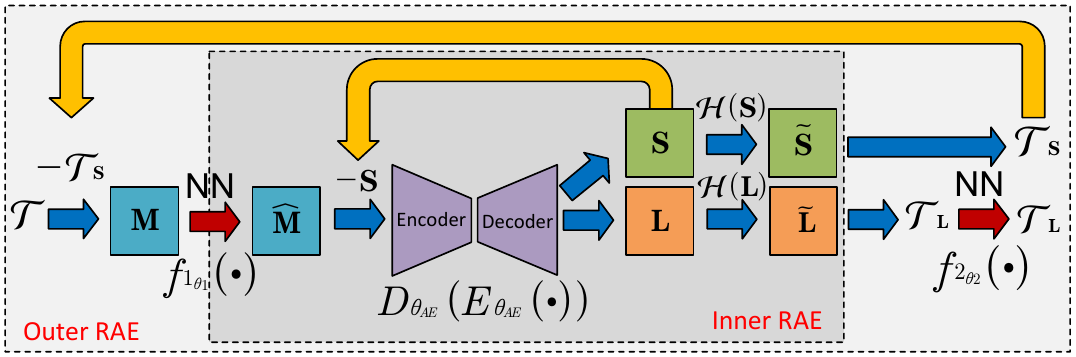}
  \captionof{figure}{\texttt{RDAE} framework.}
  \label{fig:RDAE}
\end{minipage}
\end{figure*}

\subsection{Robust Autoencoders}
\label{sec:rae}
\rev{We propose a \emph{Robust Autoencoder}~(\texttt{RAE}) that combines the benefits of an \texttt{AE} and \texttt{RPCA}, namely the non-linear computations and temporal dependencies of \texttt{AE}s and the robustness of \texttt{RPCA} (cf. Table~\ref{tab:advantages})}. 
\rev{The key insight is that the hidden representation in the bottleneck layer of an \texttt{AE} is very compact, and thus it is only possible to reconstruct representative features from the input data, but not the specifics of the input data such as outliers. %
Thus, outliers are not easily compressible and thus cannot be represented well in the bottleneck layer of an \texttt{AE}.}
Thus, we aim at obtaining a clean time series $\mathcal{T}_\mathbf{L}$ by removing an outlier time series $\mathcal{T}_\mathbf{S}$ from the original time series $\mathcal{T}$, i.e., $\mathcal{T}_\mathbf{L}=\mathcal{T}-\mathcal{T}_\mathbf{S}$. We then employ an \texttt{AE} to reconstruct only the clean time series $\mathcal{T}_\mathbf{L}$. Fig.~\ref{fig:RAE} shows the overview of \texttt{RAE} framework.
We formulate the following training objective:

\begin{equation}
	\begin{aligned}
		\argmin_{\theta_{AE}, \mathcal{T}_\mathbf{S}} \; &||\mathcal{T}_\mathbf{L} - D_{\theta_{AE}}(E_{\theta_{AE}}(\mathcal{T}_\mathbf{L}))||_2 + \lambda||\mathcal{T}_\mathbf{S}||_0 \\ \text{s.t.} \; &\mathcal{T} = \mathcal{T}_\mathbf{L} + \mathcal{T}_\mathbf{S}.
	\end{aligned}
	\label{fn:optimization_rae_0}
\end{equation}

The objective function of the \texttt{AE} in Eq.~\ref{fn:optimization_rae_0} has two terms. 
The first term measures the discrepancy between $\mathcal{T}_\mathbf{L}$ and its reconstruction  $D_{\theta_{AE}}(E_{\theta_{AE}}(\mathcal{T}_\mathbf{L}))$. 
Since $\mathcal{T}_\mathbf{L}$ is supposed to have only clean data, an \texttt{AE} should be able to construct $\mathcal{T}_\mathbf{L}$ easily. 
The second term encourages the sparsity of outlier signals $\mathcal{T}_\mathbf{S}$ via $\ell_0$ regularization, \rev{which is also used in \texttt{RPCA}~\cite{DBLP:journals/jacm/CandesLMW11}}. \rev{Here, $\ell_0$ counts the non-zero elements, which ensures the property of outliers that the number of outliers is small. Otherwise, they are not outliers, but instead represent some regular patterns~\cite{DBLP:books/sp/Aggarwal13}.} 
Hyperparameter $\lambda$ controls the relative importance of the regularization.  

Optimizing Eq.~\ref{fn:optimization_rae_0} identifies not only the optimal parameters for the \texttt{AE}, $\theta_{AE}$, but also the optimal outlier time series $\mathcal{T}_\mathbf{S}$. 
However, due to the constraint $\mathcal{T}=\mathcal{T}_\mathbf{L}+\mathcal{T}_\mathbf{S}$, traditional learning algorithms for neural networks, e.g., stochastic gradient descent, are not applicable. 
We design a bijective learning algorithm that alternates between (i) optimizing the first term to update $\theta_{AE}$ while keeping $\mathcal{T}_\mathbf{S}$ unchanged and (ii) optimizing the second term to update $\mathcal{T}_\mathbf{S}$ while keeping $\theta_{AE}$ unchanged.  
We ensure that the constraint is satisfied during the optimization. 
We detail the learning algorithm in Section~\ref{sec:training}. 

We choose a 1D convolutional neural network (1D \texttt{CNN}) as the computational units to build the \texttt{AE} because 1D \texttt{CNN}s have been shown to have an impressive performance on temporal data~\cite{DBLP:conf/mdm/Kieu0J18}. 
The 1D \texttt{CNN}-based \texttt{RAE} consists of two parts---an encoder $E_{\theta_{AE}}(\cdot)$ and a decoder $D_{\theta_{AE}}(\cdot)$, where each part consists of several 1D convolutional layers. 
The operations in the encoder $E_{\theta_{AE}}(\cdot)$ and the decode $D_{\theta_{AE}}(\cdot)$ are shown in Eqs.~\ref{fn:1dcnn_encoder} and~\ref{fn:1dcnn_decoder}.

\begin{align}
	E_{\theta_{AE}}(\mathcal{T}_\mathbf{L}) &\defeq \phi(\mathbf{W}_e \ast \mathcal{T}_\mathbf{L} + \mathbf{b}_e)
	\label{fn:1dcnn_encoder}\\
	D_{\theta_{AE}}(\mathcal{T}_\mathbf{L}) &\defeq \phi(\mathbf{W}_d \ast E_{\theta_{AE}}(\mathcal{T}_\mathbf{L}) + \mathbf{b}_d)
	\label{fn:1dcnn_decoder}
\end{align}

Here, $\phi$ is a non-linear activation function, e.g., $\mathrm{sigmoid}$, $\mathrm{relu}$, or $\mathrm{tanh}$. $\mathbf{W}_e$ and $\mathbf{b}_e$ are the weight matrix and bias vector of the encoder to create the feature maps in the encoder, respectively. 
$\mathbf{W}_d$  and $\mathbf{b}_d$ are the counterparts in the decoder. 
Generally, the number of feature maps of the encoder is less than the number of feature maps of the decoder to form a bottleneck layer.
$\ast$ denotes the 1D convolution operator.
In the encoding phase, following the convolutional layer, we use a max-pooling
(i.e., down sampling method) layer to condense the data as well as increase the effect of having a bottleneck layer.
In the decoding phase, in contrast to max-pooling, up-sampling is used to construct the output. 
The parameters in the \texttt{AE} include $\theta_{AE}=\{\mathbf{W}_e$, $\mathbf{b}_e$, $\mathbf{W}_d, \mathbf{b}_d\}$, which need to be learned.

Fig.~\ref{fig:RAE} shows an overview of \texttt{RAE}. 
$\mathcal{T}_\mathbf{S}$, which is set to be empty in the first iteration, is subtracted from an input time series $\mathcal{T}$.
The result is fed to the 1D \texttt{CNN}-based \texttt{RAE}. By solving the optimization problem shown in Eq.~\ref{fn:optimization_rae_0}, 
$\mathcal{T}$ is split into two parts,
$\mathcal{T}_\mathbf{L}$ and $\mathcal{T}_\mathbf{S}$. 
Among these, $\mathcal{T}_\mathbf{S}$ is used to update the input of the \texttt{AE}, i.e., $\mathcal{T} - \mathcal{T}_\mathbf{S}$, in the next iteration (cf.\ the yellow arrow in Fig.~\ref{fig:RAE}).

\subsection{Robust Dual Autoencoders}

\label{sec:rdae}
Multi-view learning has been shown to be able to improve the robustness of learning algorithms by providing complementary information to the learner~\cite{DBLP:conf/icml/WangALB15}. 
Intuitively, our framework could benefit from a multi-view formulation, e.g., a matrix view and a time series view.  %
Motivated by this, we propose the second framework, \emph{Robust Dual Autoencoders} (\texttt{RDAE}), which differs from \texttt{RAE} by using two robust \texttt{AE}s that help each other learn from two different representations of a time series, i.e., a matrix representation and a time series representation. 
\rev{\texttt{RDAE} also combines \texttt{AE}s and \texttt{RPCA} to leverage the benefits of both. Further, \texttt{RDAE} considers multi-view learning mechanism (cf. Table~\ref{tab:advantages}).}

Fig.~\ref{fig:RDAE} shows an overview of \texttt{RDAE}. We first embed $\mathcal{T} = \langle \mathbf{s}_1, \dots, \mathbf{s}_C \rangle$ into a lagged matrix~\cite{GG:jour/PRE/GrothG11} 
$\mathbf{M}\in \mathbb{R}^{B \times K \times D}$.
Here, we use a sliding window of size $B$, where $\displaystyle 1 < B < {C}/{2}$, to iterate through $\mathcal{T}$ to obtain $K = C - B + 1$ columns that constitute $\mathbf{M}$ as follows. 
Recall that $\mathbf{s}_i$ in the lagged matrix is a $D$-dimensional vector. The anti-diagonal elements in matrix $\mathbf{M}$ are identical; such matrices are called \emph{Hankel} matrices~\cite{DBLP:books/daglib/NinaVA01}. 

\begin{align*}
	\mathbf{M} = \begin{bmatrix} 
		\mathbf{s}_1 & \mathbf{s}_2 &  \cdots & \mathbf{s}_K \\ 
		\mathbf{s}_2 & \mathbf{s}_3 & \cdots & \mathbf{s}_{K+1}\\
		\vdots       & \vdots     & \ddots  & \vdots \\
		\mathbf{s}_B & \mathbf{s}_{B+1} & \cdots &  \mathbf{s}_{C}
	\end{bmatrix}
\end{align*}

In \texttt{RDAE}, the time series is the first representation, whereas the lagged matrix is the second representation that reflects the geometry of a time series~\cite{packard1980geometry}. More specifically, the lagged matrix is able to represent shapes and patterns of time series such as the magnitude of trends and the frequency of periodicities~\cite{packard1980geometry}. 
Instead of decomposing $\mathcal{T}$ directly, we use an inner \texttt{RAE} to decompose matrix $\mathbf{M}$ into $\mathbf{L}$ and $\mathbf{S}$. Then, based on the two matrices, we use an outer \texttt{RAE} to decompose  $\mathcal{T}$ into $\mathcal{T}_{\mathbf{L}}$ and $\mathcal{T}_{\mathbf{S}}$ (see Fig.~\ref{fig:RDAE}). 

\subsubsection{Inner RAE} 
The inner \texttt{RAE} decomposes the lagged matrix $\mathbf{M}$; this can be viewed as a nonlinear version of \texttt{RPCA}. As shown in Fig.~\ref{fig:RDAE}, rather than decomposing $\mathbf{M}$ directly, we first perform a \textit{nonlinear transformation} of $\mathbf{M}$ by applying 2D \texttt{CNN} layers $\hat{\mathbf{M}}=f_{1_{\theta_1}}({\mathbf{M}})$. %
We use 2D \texttt{CNN}s because the input is no longer a time series but a matrix. %
Specifically, we have: %

\begin{align}
	\hat{\mathbf{M}} = f_{1_{\theta_1}}(\mathbf{M}) &\defeq  \phi(\mathbf{W}_{1} \ast \mathbf{M} + \mathbf{b}_{1}) 	
	\label{fn:f_1}
\end{align}

Here, $\mathbf{W}_{1}, \mathbf{b}_{1} \in \theta_{1}$ are the weight matrices and bias vectors of the \textit{transformation}. The parameters in the transformation include $\theta_{1}=\{\mathbf{W}_1, \mathbf{b}_1\}$, which need to be learned. The output $\hat{\mathbf{M}} \in \mathbb{R}^{B \times K \times D}$ has the same size as $\mathbf{M}$. \rev{A time series often contains small variations (i.e., noise) that is not viewed as outliers, but can still affect the accuracy of the model. The transformation has the effect of smoothing $\mathbf{M}$ by removing such noise.} By doing this, we expect that the inner \texttt{RAE} achieves a cleaner decomposition, which enables a better reconstruction, thus improving robustness. We still want the smoothed matrix $\hat{\mathbf{M}}$ to be similar to ${\mathbf{M}}$. Thus, we design an objective function that minimizes the discrepancy between the two. 

\begin{align}
	\argmin_{\theta_1} ||\mathbf{M} - \hat{\mathbf{M}}||_2 = \argmin_{\theta_1} ||\mathbf{M} - f_{1_{\theta_1}}(\mathbf{M})||_2
	\label{fn:optimization_rda_0a}
\end{align}

After obtaining $\hat{\mathbf{M}}$, we employ an inner robust \texttt{AE} to decompose $\hat{\mathbf{M}}$ into $\mathbf{L}$ and $\mathbf{S}$. The \texttt{AE} has an encoder $E_{\theta_{AE}}$ and a decoder $D_{\theta_{AE}}$; both are based on 2D \texttt{CNN}s.

\begin{align}
	E_{\theta_{AE}}(\mathbf{L}) &\defeq \phi(\mathbf{W}_{e} \ast \mathbf{L} + \mathbf{b}_e) \\ 
	D_{\theta_{AE}}(\mathbf{L}) &\defeq \phi(\mathbf{W}_{d} \ast E_{\theta_{AE}}(\mathbf{L}) + \mathbf{b}_d) 
	\label{fn:autoencoder}
\end{align}

Here, $\mathbf{W}_e, \mathbf{b}_e \in \theta_{AE}$ are the weight matrices and bias vectors of the encoder, and $\mathbf{W}_d, \mathbf{b}_d \in \theta_{AE}$ are the the weight matrices and bias vectors of the decoder. The parameters in the encoder and the decoder include $\theta_{AE}=\{\mathbf{W}_e, \mathbf{b}_e, \mathbf{W}_d, \mathbf{b}_d\}$, which need to be learned.

We define a loss function for the inner \texttt{RAE} as follows.

\begin{equation}
	\begin{aligned}
		\argmin_{\theta_{AE}, \mathbf{S}} \; &||\mathbf{L} - D_{\theta_{AE}}(E_{\theta_{AE}}(\mathbf{L}))||_2 + \lambda_1||\mathbf{S}||_0 \\ \text{s.t.} \; &\hat{\mathbf{M}} = \mathbf{L} + \mathbf{S}.
	\end{aligned}
	\label{fn:optimization_rda_0b}
\end{equation}

The first term measures the discrepancy between $\mathbf{L}$ and the reconstructed $D_{\theta_{AE}}(E_{\theta_{AE}}(\mathbf{L}))$. The \texttt{AE} uses a bottleneck layer to reconstruct $\mathbf{L}$, which mimics the way \texttt{RPCA} uses a low-rank matrix. Second, it employs an $\ell_0$ regularization to $\mathcal{T}_\mathbf{S}$ following the regularization term of \texttt{RPCA}. Finally, the optimization is constrained by $\hat{\mathbf{M}} = \mathbf{L} + \mathbf{S}$. 

\subsubsection{Outer RAE} 
The outer \texttt{RAE} decomposes $\mathcal{T}$ into $\mathcal{T}_\mathbf{L}$ and $\mathcal{T}_\mathbf{S}$ while considering the output matrices $\mathbf{L}$ and $\mathbf{S}$ from the inner \texttt{RAE}. Specifically, we first employ the Hankelization operator~\cite{DBLP:books/daglib/NinaVA01} to transform matrices $\mathbf{L}$ and $\mathbf{S}$ into Hankel matrices: $\tilde{\mathbf{L}} = \mathcal{H}(\mathbf{L})$ and $\tilde{\mathbf{S}} = \mathcal{H}(\mathbf{S})$. Next, we covert the Hankel matrices back to time series $\mathcal{T}_\mathbf{L}$ an $\mathcal{T}_\mathbf{S}$ using the reverse of the operation that embeds a time series into a lagged matrix. Then we feed $\mathcal{T}_\mathbf{L}$ to 1D \texttt{CNN} layers, $f_{2_{\theta_2}}(\cdot)$, to perform a nonlinear transformation. 
The output is an updated clean time series with the same size. Specifically, we have: 

\begin{align}
	\mathcal{T}_\mathbf{L} = f_{2_{\theta_2}}(\mathcal{T}_\mathbf{L}) &\defeq  \phi(\mathbf{W}_{2} \ast \mathcal{T}_\mathbf{L} + \mathbf{b}_{2})
	\label{fn:f_2}
\end{align}

Here, $\mathbf{W}_2, \mathbf{b}_2 \in \theta_{2}$ are the weight matrices and bias vectors of the outer \texttt{RAE}. The parameters in the outer \texttt{RAE} include $\theta_2=\{\mathbf{W}_2, \mathbf{b}_2\}$, which need to be learned.
While doing this, we utilize the 1D \texttt{CNN} layers as the outer \texttt{RAE} by solving the following optimization problem. 

\begin{equation}
	\begin{aligned}
		\argmin_{\theta_{2}, \mathcal{T}_\mathbf{S}} \; &||\mathcal{T}_\mathbf{{\mathbf{L}}} - f_{2_{\theta_2}}(\mathcal{T}_\mathbf{{\mathbf{L}}})||_2 + \lambda_2||\mathcal{T}_\mathbf{S}||_0 \\ \text{s.t.} \; & \mathcal{T} = \mathcal{T}_\mathbf{L} + \mathcal{T}_\mathbf{S} 
	\end{aligned}
	\label{fn:optimization_rda_0c}
\end{equation}

This loss function (i) minimizes the loss between time series $\mathcal{T}_\mathbf{{\mathbf{L}}}$, which is derived from the clean matrix $\mathbf{L}$, and the reconstructed time series $f_{2_{\theta_2}}(\mathcal{T}_\mathbf{L})$; and (ii) it causes $\mathcal{T}_\mathbf{S}$  to be sparse due to the $\ell_0$ regularization. The optimization is constrained by $\mathcal{T} = \mathcal{T}_\mathbf{L} + \mathcal{T}_\mathbf{S}$ as the outer \texttt{RAE} decomposes the time series $\mathcal{T}$. 

\subsubsection{Interplay between the Inner and Outer RAEs} 
The two \texttt{AE}s help each other achieve the best separation of outliers using clean data from different representations. The first is based on the lagged matrix, while the second works on the time series while taking into account the lagged matrix decomposition. Having obtained $\mathcal{T}_\mathbf{L}$ from the outer \texttt{RAE}, we use $\mathcal{T}_\mathbf{L}$ as a new time series $\mathcal{T}$ such that the process continues iteratively until convergence. 

\subsection{Outlier Scoring}
Following existing studies~\cite{DBLP:conf/iccv/XiaCWHS15,DBLP:conf/mdm/Kieu0J18}, 
we assign an outlier score to each observation in time series $\mathcal{T}$. Since we decomposed $\mathcal{T}$ into a clean time series $\mathcal{T}_{\mathbf{L}}$ and an outlier time series $\mathcal{T}_{\mathbf{S}}$, the outlier scores $\mathcal{OS}$ are the $\mathrm{norm}$ values of observations in $\mathcal{T}_\mathbf{S}=\langle \mathbf{s}_{\mathbf{S}_{1}}, \mathbf{s}_{\mathbf{S}_{2}}, \dots, \mathbf{s}_{\mathbf{S}_{C}} \rangle$: 

\begin{align}
   \mathcal{OS} = \langle ||\mathbf{s}_{\mathbf{S}_{1}}||^2_2, ||\mathbf{s}_{\mathbf{S}_{2}}||^2_2, \dots, ||\mathbf{s}_{\mathbf{S}_{C}}||^2_2 \rangle 
\end{align}

We use the $\ell_2$ norm to convert vectors to scalar values that are more intuitive.

\subsection{Reformulating the Loss Functions}

\rev{The loss functions of both \texttt{RAE} and \texttt{RDAE} include a term using the $\ell_0$ norm that aims to make the outlier matrix or outlier time series sparse, thus ensuring the semantics of outliers. However, the $\ell_0$ norm is non-convex, which makes optimization difficult~\cite{DBLP:journals/tit/CandesT05}. 
Following the literature~\cite{DBLP:journals/mp/GeJY11}, we therefore relax the $\ell_0$ norm to the $\ell_1$ norm that is a good approximation of the $\ell_0$ norm.} 
The reformulated loss function of the \texttt{RAE} is shown in Eq.~\ref{fn:optimization_rae_0_l1}.

\begin{equation}
	\begin{aligned}
		\argmin_{\theta_{AE}, \mathcal{T}_\mathbf{S}} \; &||\mathcal{T}_\mathbf{L} - D_{\theta_{AE}}(E_{\theta_{AE}}(\mathcal{T}_\mathbf{L}))||_2 + \lambda||\mathcal{T}_\mathbf{S}||_1 \\ \text{s.t.} \;  &\mathcal{T} = \mathcal{T}_\mathbf{L} + \mathcal{T}_\mathbf{S}
	\end{aligned}
	\label{fn:optimization_rae_0_l1}
\end{equation}

The reformulated loss functions of \texttt{RDAE} are shown in Eqs.~\ref{fn:optimization_rda_1a_l1}, \ref{fn:optimization_rda_1b_l1}, and \ref{fn:optimization_rda_1c_l1}.

\begin{numcases}{}
	\begin{aligned}
		\argmin_{\theta_1} \; &||\mathbf{M} - f_{1_{\theta_1}}(\mathbf{M})||_2
		\label{fn:optimization_rda_1a_l1}
	\end{aligned}
	\\
	\begin{aligned}
		\argmin_{\theta_{AE}, \mathbf{S}} \; &||\mathbf{L} - D_{\theta_{AE}}(E_{\theta_{AE}}(\mathbf{L}))||_2 + \lambda_1||\mathbf{S}||_1 \\ \text{s.t.} \; &\hat{\mathbf{M}} = \mathbf{L} + \mathbf{S}
		\label{fn:optimization_rda_1b_l1}
	\end{aligned}
	\\
	\begin{aligned}
		\argmin_{\theta_{2}, \mathcal{T}_\mathbf{S}} \; &||\mathcal{T}_\mathbf{L} - f_{2_{\theta_{2}}}(\mathcal{T}_\mathbf{L})||_2 + \lambda_2||\mathcal{T}_\mathbf{S}||_1 \\
		\text{s.t.} \; &\mathcal{T} = \mathcal{T}_\mathbf{L} + \mathcal{T}_\mathbf{S} 
		\label{fn:optimization_rda_1c_l1}
	\end{aligned}
	\label{fn:optimization_rda_1}
\end{numcases}

\subsection{Training Algorithms}
\label{sec:training}
The optimization problems of \texttt{RAE} and \texttt{RDAE} have constraints and thus cannot be solved by gradient descent based back-propagation (\texttt{BACKPROP}). %
Our optimization problems are two-block optimization problems, which can be solved by the \emph{Alternating Direction Method of Multipliers} (\texttt{ADMM})~\cite{DBLP:journals/fa/BoydNPEJ11}. The core idea of \texttt{ADMM} is to divide an objective function into multiple sub-objectives. Then, \texttt{ADMM} optimizes one sub-objective function while keeping the remaining sub-objectives fixed. After optimizing a sub-objective function, the constraint is enforced. In addition, we use the \emph{Proximal Algorithm} (\texttt{PROX})~\cite{DBLP:journals/fa/NealB14} to optimize elements inside the $\ell_1$ norm.

\subsubsection{Optimizing RAE (Algorithm 1)} 

\begin{algorithm}[h]
	\KwInput{Time series $\mathcal{T}$, double $\lambda$, double $\epsilon$}
	\KwOutput{$\mathcal{T}_\mathbf{L}, \mathcal{T}_\mathbf{S}$}
	Initialization: $\mathcal{T}_\mathbf{L} \leftarrow 0$; $\mathcal{T}_\mathbf{S} \leftarrow 0$; $\mathcal{T}^{*} \leftarrow \mathcal{T}$;\\
	\Repeat{$condition_1 < \epsilon$ or $condition_2 < \epsilon$}
	{	
		$\mathcal{T}_\mathbf{L} \leftarrow \mathcal{T} - \mathcal{T}_\mathbf{S}$;\\
		\tcp{Optimize \texttt{AE} $\theta_{AE}$}
		Update $\theta_{AE}$ by minimizing $||\mathcal{T}_\mathbf{L} - D_{\theta_{AE}}(E_{\theta_{AE}}(\mathcal{T}_\mathbf{L}))||_2$ using \texttt{BACKPROP};\\
		$\mathcal{T}_\mathbf{L} \leftarrow D_{\theta_{AE}}(E_{\theta_{AE}}(\mathcal{T}_\mathbf{L}))$;\\
		$\mathcal{T}_\mathbf{S} \leftarrow \mathcal{T} - \mathcal{T}_\mathbf{L}$;\\
		\tcp{Optimize $\mathcal{T}_\mathbf{S}$}
		Update $\mathcal{T}_\mathbf{S}$ by minimizing $\lambda||\mathcal{T}_\mathbf{S}||_1$ using \texttt{PROX};\\
		\tcp{Compute stopping conditions}
		$\displaystyle condition_1 \leftarrow {||\mathcal{T} - \mathcal{T}_\mathbf{L} - \mathcal{T}_\mathbf{S}||_2}/{||\mathcal{T}||_2}$;\\
		$\displaystyle condition_2 \leftarrow {||\mathcal{T}^{*} - \mathcal{T}_\mathbf{L} - \mathcal{T}_\mathbf{S}||_2}/{||\mathcal{T}||_2}$;\\
		$\mathcal{T}^{*} \leftarrow \mathcal{T}_\mathbf{L} + \mathcal{T}_\mathbf{S}$;\\
	}	
	\Return $\mathcal{T}_\mathbf{L}, \mathcal{T}_\mathbf{S}$;\\
	\caption{Training algorithm for \texttt{RAE}}
	\label{al:training_rae}
\end{algorithm}

We split the reformulated objective function of \texttt{RAE} (see Eq.~\ref{fn:optimization_rae_0_l1}) into two parts. 
The first part optimizes $\theta_{AE}$ to minimize $||\mathcal{T}_\mathbf{L} - D_{\theta_{AE}}(E_{\theta_{AE}}(\mathcal{T}_\mathbf{L}))||_2$; the second part optimizes $\mathcal{T}_\mathbf{S}$ to minimize $\lambda||\mathcal{T}_\mathbf{S}||_1$. 
Finally, we update $\mathcal{T}_\mathbf{L}$ to be $\mathcal{T} - \mathcal{T}_\mathbf{S}$ to enforce the constraint. 
The result is then fed into the next iteration (see the yellow arrow in Fig.~\ref{fig:RAE}). 
The first stopping condition, \textit{condition}$_1$, ends the optimization when it satisfies the constraints $\mathcal{T} = \mathcal{T}_\mathbf{L} + \mathcal{T}_\mathbf{S}$. 
The second stopping condition, \textit{condition}$_2$, ends the optimization when $\mathcal{T}_\mathbf{L}$ and $\mathcal{T}_\mathbf{S}$ are fixed.

\subsubsection{Optimizing RDAE (Algorithm 2)} 

\begin{algorithm}[t]
	\KwInput{Time series $\mathcal{T}$, window length $B$, double $\lambda_1$, double $\lambda_2$, double $\epsilon$}
	\KwOutput{$\mathcal{T}_\mathbf{L}, \mathcal{T}_\mathbf{S}$}
	Initialization: $\mathbf{L} \leftarrow 0$; $\mathbf{S} \leftarrow 0$; $\mathcal{T}_\mathbf{L} \leftarrow 0$; $\mathcal{T}_\mathbf{S} \leftarrow 0$; \\
	\While{not converged}
	{
	    $\mathcal{T}_{\mathbf{L}} \leftarrow \mathcal{T} - \mathcal{T}_{\mathbf{S}}$;\\
		Create lagged matrix $\mathbf{M} \in \mathbb{R}^{B \times K \times D}$ from $\mathcal{T}_{\mathbf{L}}$;\\
		Update $\theta_1$ by minimizing $||\mathbf{M} - f_{1_{\theta_1}}(\mathbf{M})||_2$ using \texttt{BACKPROP}; \\ 
		$\hat{\mathbf{M}} \leftarrow f_{1_{\theta_1}}(\mathbf{M})$;\\
		$\hat{\mathbf{M}}^{*} \leftarrow \hat{\mathbf{M}}$;\\
		\Repeat{$condition_1 < \epsilon$ or $condition_2 < \epsilon$} 
		{	
			$\mathbf{L} \leftarrow \hat{\mathbf{M}} - \mathbf{S}$;\\
			\tcp{Optimize inner \texttt{AE} $\theta_{AE}$}
			Update $\theta_{AE}$ by minimizing $|| \mathbf{L} - D_{\theta_{AE}}(E_{\theta_{AE}}(\mathbf{L})) ||_2$ using \texttt{BACKPROP};\\
			$\mathbf{L} \leftarrow D_{\theta_{AE}}(E_{\theta_{AE}}(\mathbf{L}))$;\\
			$\mathbf{S} \leftarrow \hat{\mathbf{M}} - \mathbf{L}$;\\
			\tcp{Optimize $\mathbf{S}$}
			Update $\mathbf{S}$ by minimizing $\lambda_1||\mathbf{S}||_1$ using \texttt{PROX};  \\
			\tcp{Compute stopping conditions}
            $\displaystyle condition_1 \leftarrow {||\hat{\mathbf{M}} - \mathbf{L} - \mathbf{S}||_2}/{||\hat{\mathbf{M}}||_2}$;\\
			$\displaystyle condition_2 \leftarrow {||\hat{\mathbf{M}}^{*} - \mathbf{L} - \mathbf{S}||_2}/{||\hat{\mathbf{M}}||_2}$;\\
			$\hat{\mathbf{M}}^{*} \leftarrow \mathbf{L} + \mathbf{S}$;\\
		}	
		Create Hankel matrices $\tilde{\mathbf{L}} \leftarrow \mathcal{H}(\mathbf{L})$ and
		$\tilde{\mathbf{S}} \leftarrow \mathcal{H}(\mathbf{S})$; \\
		Obtain time series $\mathcal{T}_\mathbf{L}$ and $\mathcal{T}_\mathbf{S}$ from the two Hankel matrices; \\
		Initialization: $\mathcal{T}^{*} \leftarrow \mathcal{T}$;\\
		\Repeat{$condition_1 < \epsilon$ or $condition_2 < \epsilon$}
		{	
			$\mathcal{T}_\mathbf{L} \leftarrow \mathcal{T} - \mathcal{T}_\mathbf{S}$;\\
			\tcp{Optimize outer \texttt{AE} $\theta_2$}
			Update $\theta_2$ by minimizing $|| \mathcal{T}_\mathbf{L} - f_{2_{\theta_2}}(\mathcal{T}_\mathbf{L})) ||_2$ using \texttt{BACKPROP}; \\
			$\mathcal{T}_\mathbf{L} \leftarrow f_{2_{\theta_2}}(\mathcal{T}_\mathbf{L}))$;\\
			$\mathcal{T}_\mathbf{S} \leftarrow \mathcal{T} - \mathcal{T}_\mathbf{L}$;\\
			\tcp{Optimize $\mathcal{T}_\mathbf{S}$}
			Update $\mathcal{T}_\mathbf{S}$ by minimizing $\lambda_2||\mathcal{T}_\mathbf{S}||_1$ using \texttt{PROX};\\
			\tcp{Compute stopping conditions}
            $\displaystyle condition_1 \leftarrow {||\mathcal{T} - \mathcal{T}_\mathbf{L} - \mathcal{T}_\mathbf{S}||_2}/{||\mathcal{T}||_2}$;\\
			$\displaystyle condition_2 \leftarrow {||\mathcal{T}^{*} - \mathcal{T}_\mathbf{L} - \mathcal{T}_\mathbf{S}||_2}/{||\mathcal{T}||_2}$;\\
			$\mathcal{T}^{*} \leftarrow \mathcal{T}_\mathbf{L} + \mathcal{T}_\mathbf{S}$;\\ 
		}
	}
	\Return $\mathcal{T}_{\mathbf{L}}, \mathcal{T}_{\mathbf{S}}$;\\
	\caption{Training algorithm for \texttt{RDAE}}
	\label{al:training_rdae}
\end{algorithm}

We use a loop to control the overall flow. Inside the loop, we optimize each reformulated objective function of \texttt{RDAE} in turn (see Eqs.~\ref{fn:optimization_rda_1a_l1}, \ref{fn:optimization_rda_1b_l1}, and \ref{fn:optimization_rda_1c_l1}). For Eq.~\ref{fn:optimization_rda_1a_l1}, we optimize for $\theta_1$ to minimize $||\mathbf{M} - f_{1_{\theta_1}}(\mathbf{M})||_2$ using \texttt{BACKPROP}. 
For Eq.~\ref{fn:optimization_rda_1b_l1}, we split the optimization problem into two parts. 
The first part trains the inner \texttt{RAE} by finding an optimal $\theta_{AE}$ to minimize $||\mathbf{L} - D_{\theta_{AE}}(E_{\theta_{AE}}(\mathbf{L}))||_2$; the second part finds an optimal $\mathbf{S}$ to minimize $\lambda_1||\mathbf{S}||_1$. We also need to ensure that the constraint $\hat{\mathbf{M}} = \mathbf{L} + \mathbf{S}$ is satisfied.
For Eq.~\ref{fn:optimization_rda_1c_l1}, we also split the optimization problem into two parts. 
The first part trains the outer \texttt{RAE} by finding an optimal $\theta_2$ to minimize $||\mathcal{T}_\mathbf{L} - f_{2_{\theta_{2}}}(\mathcal{T}_\mathbf{L})||_2$; 
the second part finds an optimal $\mathcal{T}_\mathbf{S}$ to minimize $\lambda_2||\mathcal{T}_\mathbf{S}||_1$. 
\rev{To connect the inner \texttt{RAE} and the outer \texttt{RAE}, which are trained by \texttt{ADMM}, we subtract $\mathcal{T}_\mathbf{S}$ from $\mathcal{T}$ to update $\mathbf{M}$ and $\hat{\mathbf{M}}$ (lines 4 and 6). Then, $\mathbf{S}$ is subtracted from $\hat{\mathbf{M}}$ (line 9) to update $\mathbf{L}$ (line 11) and $\mathbf{S}$ (line 13). 
From the updated $\mathbf{L}$ and $\mathbf{S}$ (line 18), $\mathcal{T}_\mathbf{L}$ and $\mathcal{T}_\mathbf{s}$ are updated (lines 24 and 26).
When subtracting $\mathcal{T}_\mathbf{S}$ from $\mathcal{T}$ in the first few iterations, $\mathcal{T}_\mathbf{S}$ at that time has not yet converged. Then, the $\hat{\mathbf{M}}$ has not converged either. Thus, $\mathbf{S}$ must be subtracted from $\hat{\mathbf{M}}$.}
The first stopping condition in the inner \texttt{RAE} and in the outer \texttt{RAE}, \textit{condition}$_1$, ends the optimization when it satisfies the constraint $\hat{\mathbf{M}} = \mathbf{L} + \mathbf{S}$ and the constraints $\mathcal{T} = \mathcal{T}_\mathbf{L} + \mathcal{T}_\mathbf{S}$, respectively. 
The second stopping condition in the inner \texttt{RAE} and in the outer \texttt{RAE}, \textit{condition}$_2$, ends the optimization when $\mathbf{L}$ and $\mathbf{S}$ are fixed and when $\mathcal{T}_\mathbf{L}$ and $\mathcal{T}_\mathbf{S}$ are fixed, respectively.

\subsubsection{Convergence Analysis}
\label{ssec:convergence}
Although the objective functions used in \texttt{RAE} and \texttt{RDAE} are not convex, %
following existing studies~\cite{DBLP:conf/icml/NishiharaLRPJ15,DBLP:journals/siamjo/HongLR16}, optimizing them using \texttt{ADMM} ensures convergence~\cite{DBLP:conf/icml/NishiharaLRPJ15,DBLP:journals/siamjo/HongLR16}. 
In addition, we provide empirical evidence for the convergence of the proposed algorithms in Section~\ref{ssec:empirical_convergence_analysis}. 

\subsubsection{\rev{Complexity Analysis}}
\label{ssec:complexity}
\rev{The complexity of \texttt{RAE} is $O(e \cdot C \cdot D \cdot N \cdot L)$, where $e$ is the number of epochs, $C$ is the time series length, $D$ is the time series dimensionality, $N$ is the number of kernels, and $L$ is number of convolutional layers. 
The complexity of \texttt{RDAE} is the sum of the complexities of the transformation, the inner \texttt{RAE}, and the outer \texttt{RAE}. Here, the inner \texttt{RAE}'s complexity dominates the other complexities because the inner \texttt{RAE} works on the lagged matrix that is much larger than the original time series. The complexity of the inner \texttt{RAE} is $O(e \cdot B \cdot K \cdot D \cdot N \cdot L)$, where $B$ is the window size and $K=C-B+1$.}

\section{Explainability Analysis}
\label{sec:explainability}
\subsection{Concept and Motivation}
\rev{Autoencoder based outlier detection methods regard observations with large reconstruction errors, i.e., large deviations from the ``clean'' time series, as outliers.} 
In traditional \texttt{AE}s, the reconstructed time series are considered as the clean time series. In the two proposed robust autoencoder frameworks, the decomposed clean time series $\mathcal{T}_\mathbf{L}$ is the clean time series. 
The clean time series represents the observations that should occur if the underlying system generating the time series is in a normal state; and the observations with large deviations w.r.t. the corresponding observations from the clean time series, i.e., 
with large reconstruction errors, are outliers. %
Thus, to explain why an autoencoder based method regards some observations as outliers, it is important to understand the reconstruction errors $\mathcal{T}-\mathcal{T}_\mathbf{L}$. Since the input time series $\mathcal{T}$ is given, it is then important that the users understand the key features of the clean time series $\mathcal{T}_\mathbf{L}$ returned by the autoencoder method, e.g., increasing or decreasing trends, and different periodicities. This helps users know which observations should occur in a normal state, which then facilitates the users to understand the reconstruction errors and the identified outliers.

Based on the above intuition, \rev{we propose two explainability scores to quantify the explainability of different autoencoder based outlier detection methods by measuring how easily a user can understand the key features of the clean time series (e.g., trend and seasonality) returned by such methods.} We expect a framework to offer high explainability, if its derived clean time series includes easy-to-understand patterns (e.g., trend and seasonality). Then a user can easily grasp the key features of the clean time series and understand which observations should occur at particular timestamps. 
The easier a user can understand the patterns in the clean time series, the higher the explainability of a framework. 

\begin{figure*}[htb]
    \vspace{-1.0em}
    \begin{subfigure}[t]{0.24\linewidth}
		\includegraphics[width=1.0\linewidth]{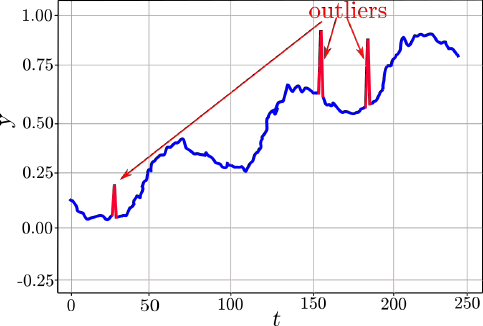}
		\caption{Input time series $\mathcal{T}$}%
		\label{subfig:explainability_a}
	\end{subfigure}
	\begin{subfigure}[t]{0.24\linewidth}
		\includegraphics[width=1.0\linewidth]{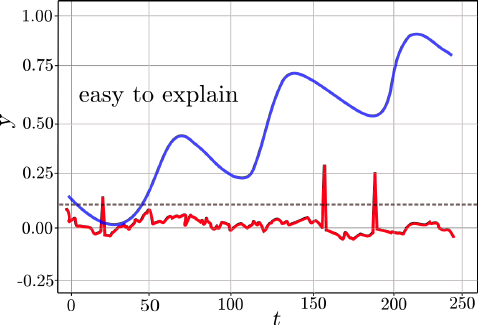}
		\caption{Framework $A$, high explainability, high accuracy}%
		\label{subfig:explainability_b}
	\end{subfigure}
	\begin{subfigure}[t]{0.24\linewidth}
		\includegraphics[width=1.0\linewidth]{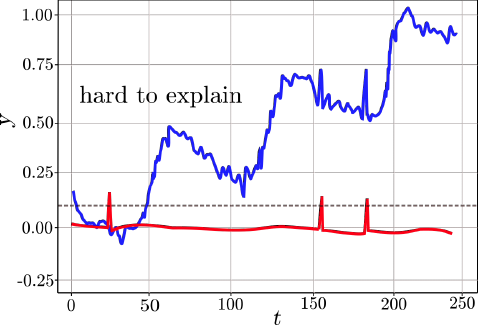}
		\caption{Framework $B$, low explainability, high accuracy}%
		\label{subfig:explainability_c}
	\end{subfigure}
		\begin{subfigure}[t]{0.24\linewidth}
		\includegraphics[width=1.0\linewidth]{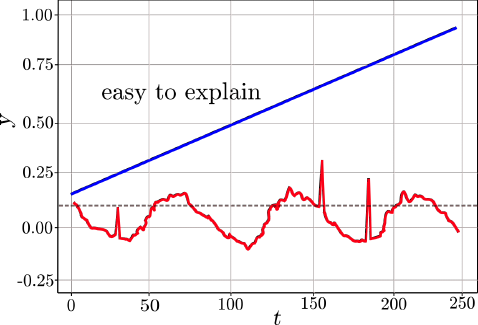}
		\caption{Framework $C$, high explainability, low accuracy}%
		\label{subfig:explainability_d}
	\end{subfigure}
	\caption{Intuition of explainability:
	in (b), (c), and (d), the blue curves show the clean time series  $\mathcal{T}_{\mathbf{L}}^{A}$, $\mathcal{T}_{\mathbf{L}}^{B}$, and $\mathcal{T}_{\mathbf{L}}^{C}$, and the red curves show the outlier time series, obtained from frameworks $A$, $B$, and $C$, respectively. The dashed horizontal line represents the outlier score threshold 0.15.
	Frameworks $A$ and $B$ both find the three outliers as their outlier time series have three clear peaks which are above the horizontal line, while framework $C$ is very inaccurate in terms of detecting outliers.
	Time series $\mathcal{T}_{\mathbf{L}}^{A}$ and $\mathcal{T}_{\mathbf{L}}^{C}$ can be explained well using trend and periodicity, thus having high explainability, while $\mathcal{T}_{\mathbf{L}}^{B}$ cannot be explained well because it includes many hard-to-explain variations, thus having low explainability. Framework $A$ is highly desirable as it has both high accuracy and high explainability. 
    }
	\label{fig:explainability}
\end{figure*}

Fig.~\ref{fig:explainability} illustrates the intuition. 
It shows an input time series $\mathcal{T}$ with three outliers as well as three autoencoder methods $A$, $B$, and $C$ that each derives a clean time series (in blue) to represent the input time series, denoted by $\mathcal{T}_\mathbf{L}^{A}$, $\mathcal{T}_\mathbf{L}^{B}$, and $\mathcal{T}_\mathbf{L}^{C}$, respectively.
We consider framework $A$ to be more explainable than framework $B$ because $\mathcal{T}_\mathbf{L}^{A}$ captures the monotonically increasing trend and the periodicities of the input time series. 
Users can easily grasp the key features of the clean time series and thus understand which observations should occur at different timestamps.
In contrast, the clean time series $\mathcal{T}_\mathbf{L}^{B}$ produced by framework $B$ is more complex and includes arbitrary variations that cannot easily be attributed to the underlying process being sampled, thus making it difficult for a user to understand the observations at particular timestamps.

If we want to represent the clean time series derived by methods $A$ and $B$ by functions, we have to use a more complex function to represent accurately time series $\mathcal{T}_\mathbf{L}^{B}$ than to represent time series $\mathcal{T}_\mathbf{L}^{A}$. 
This indicates that framework $B$ has a lower explainbility than framework $A$. 
We propose two explainability metrics %
that build on Polynomial Regression Models (\texttt{PRM})~\cite{DBLP:books/sp/HastieFT01} and Singular Spectrum Analysis (\texttt{SSA})~\cite{GG:jour/PRE/GrothG11} to enable quantification of explainablity. %

Finally, framework $C$ in Fig.~\ref{fig:explainability} is even more explainable than framework $A$---time series $\mathcal{T}_\mathbf{L}^{C}$ captures only the monotonically increasing trend of the input time series and thus can be described accurately by an even simpler function. 
However, this occurs at the cost of poor \emph{outlier detection accuracy}, as many observations have large outlier scores and are thus mistakenly considered as outliers. 
We include the third framework $C$ to make the point that a solution cannot just produce very simple clean time series that give high explainability---methods are constrained by also having to be accurate in terms of outlier detection. 
In the experiments, we show that the two proposed robust methods achieve high explainability while maintaining very high accuracy.

The proposed explainability analysis along with the corresponding quantification method differs from identifying the ``root causes" of outliers~\cite{DBLP:conf/icdm/YehZUBDDSMK16,DBLP:conf/edbt/ZhangDM17,DBLP:conf/debs/RadSJD21}. 
\rev{The root cause of outliers are application/domain specific} and indicate the reasons for malfunctions in the underlying system that generates the time series~\cite{DBLP:conf/icdm/YehZUBDDSMK16}. 
\rev{For example, a root cause can be an unexpected behavior (e.g., a network flooding attack in server-load time series), an accidental event (e.g., a car accident in traffic flow time series), an external impact (e.g., an international conflict causes a dramatic increase in energy prices), or a system failure (e.g., a short circuit in time series that monitor an electrical system).} 
Identifying such root causes relies on supervised learning methods, e.g., classification methods \rev{that assign predefined labels that associate different root causes with identified outliers}. 
This requires human experts to annotate manually a large amount of observations, indicating both which observations are outliers and the corresponding root causes of the outliers.

\rev{A few application/domain-agnostic root cause analysis methods exist~\cite{DBLP:conf/debs/RadSJD21}, which work in unsupervised settings. 
Such methods identify the most anomalous channel for each detected outlier observation.} 
\rev{For example, a peak in a CPU-load channel can cause an outlier in a 3-dimensional server-load time series that consists of CPU-, RAM-, and Network-load channels. Here, the CPU-load channel is considered as the root cause for the outlier.}

\rev{In summary, the existing root cause identification methods try to explain why specific observations are identified as outliers.}
\rev{Unlike the existing methods, %
we aim at quantifying the explainability of autoencoder based outlier detection methods rather than explaining individual outlier observations by identifying their underlying causes. 
We use the explainability scores to evaluate which autoencoder based outlier detection methods are more explainable, rather than explaining individual outliers.  
Our explainability analysis and root cause identification serve different purposes and are orthogonal. Thus, they are incomparable. }

Next, we proceed to introduce the two metrics that quantify the explainbility scores. 

\subsection{PRM-based Explainability Scores} 
We view a clean time series $\mathcal{T}_{\mathbf{L}}=\langle \mathbf{y}_1$, $\mathbf{y}_2$, $\dots$,  $\mathbf{y}_C \rangle$ as consisting of $C$ training instances. We then employ a Linear Regression framework that considers $N+1$ independent variables $\{t^n\}_{n=0}^N$ to fit the training instances. When $N=1$, we fit the training instances to a linear function to capture the linear trend of the clean time series; when $N>1$, we fit the training instances using a high-order polynomial function that embodies both linear trends and nonlinear elements. 

\rev{For a given $N$, we are able to obtain the best fitting polynomial function, denoted as $\mathcal{T}_{\mathit{PRM}}^{(N)}$.} We then compute the root mean squared error $\mathit{RMSE}(\mathcal{T}_{\mathbf{L}},\mathcal{T}_{\mathit{PRM}}^{(N)})$ to measure the fit between $\mathcal{T}_{\mathit{PRM}}^{(N)}$ and the clean time series ${\mathcal{T}_{\mathbf{L}}}$. 
We define the explainability score $\mathcal{ES}_\mathit{PRM}$ as the smallest $N$ such that the $\mathit{RMSE}$ is below a given threshold $\gamma$. 
\rev{In other words, given an \textit{RMSE} threshold $\gamma$, we aim to find the smallest $N$ such that the \textit{RMSE} between the reconstructed, clean time series $\mathcal{T}_\mathbf{L}$ from the autoencoders and the time series $\mathcal{T}_{\mathit{PRM}}^{(N)}$ approximated by \textit{PRM} with order $N$ is below $\gamma$.}

\begin{align}
	\mathcal{ES}_\mathit{PRM} = \min \{N \in \mathbb{N}|\mathit{RMSE}(\mathcal{T}_{\mathbf{L}}, \mathcal{T}_{\mathit{PRM}}^{(N)})<\gamma \}
	\label{fn:explain_prm}
\end{align}
Here, $\mathbb{N}$ is the natural numbers. For a specific $\gamma$, a smaller explainability score suggests higher explainability. A smaller explainability score implies that it is possible to use a lower-order polynomial function, i.e., a simpler function, to fit the clean time series (within the same error threshold $\gamma$), which is easier for users to understand. \rev{The complexity of \texttt{PRM} is $O(C^2 \cdot D)$, where $C$ is the time series length and $D$ is the time series dimensionality.} %

\subsection{SSA-based Explainability Scores}
\texttt{SSA}~\cite{GG:jour/PRE/GrothG11} is an explainable time series analysis method that decomposes a time series into multiple components with different importance levels. These components indicate a \emph{trend} (i.e., the most important component), multiple \emph{periodicities} (i.e., less important components), and \emph{noise} (i.e., the least important components). We apply \texttt{SSA} to a clean time series $\mathcal{T}_{\mathbf{L}}$ to obtain these components. We then construct time series \rev{$\mathcal{T}_{\mathit{SSA}}^{(N)}$} by combining the top-$N$ most important components and apply $\mathit{RMSE}$ to measure the fit between \rev{$\mathcal{T}_{\mathit{SSA}}^{(N)}$} and $\mathcal{T}_{\mathbf{L}}$. We again define the explainability score $\mathcal{ES}_\mathit{SSA}$ as the smallest $N$ such that the $\mathit{RMSE}$ is below a given threshold $\gamma$.

\begin{align}
	\mathcal{ES}_\mathit{SSA} = \min \{N \in \mathbb{N}|\mathit{RMSE}(\mathcal{T}_{\mathbf{L}}, \mathcal{T}_{\mathit{SSA}}^{(N)})<\gamma \}
	\label{fn:explain_ssa}
\end{align}

Given a specific $\gamma$, a lower explainability score indicates that it is possible to use fewer components to fit the clean time series, which indicates that the clean time series is easier to understand. \rev{The complexity of \texttt{SSA} is $O(D \cdot \mathrm{min}(B^{3}, K^{3}))$, where $D$ is the time series dimensionality, $B$ is the window size, and $K=C-B+1$.}
\section{Experiments}
\label{sec:experiments}
\subsection{Experimental Setup}
\subsubsection{Datasets} 
We use five public, real-world time series datasets:
    (1) \textit{GD}\footnote{\label{data:gd}https://kaggle.com/init-owl/genesis-demonstrator-data-for-machine-learning} contains two 20-dimensional time series and three 24-dimensional time series, which are collected from pick-and-place robots. Each time series contains from around 6,000 to 16,000 observations;
    (2) \textit{HSS}\footnote{\label{data:hss}https://kaggle.com/init-owl/high-storage-system-data-for-energy-optimization} contains four 20-dimensional time series that are collected from a high storage system. Each time series contains from 19,000 to 25,000 observations;
    (3) \textit{ECG}\footnote{\label{data:ecgtd}https://cs.ucr.edu/$\sim$eamonn/discords/} contains 2-dimensional electrocardiogram time series collected from seven patients, each with 3,750 to 5,400 observations;
    (4) \textit{NAB}\footnote{\label{data:nab}https://github.com/numenta/nab/} contains time series from six domains: urban traffic, temperature, CPU workload, Twitter posts, and exchange rates. Each domain has approximately 10 time series, each with 5,000 to 20,000 observations;
    (5) \textit{S5}\footnote{\label{data:s5}https://webscope.sandbox.yahoo.com/} includes a synthetic and a real-world dataset representing the workloads of different Yahoo services. Each dataset has around 100 time series, each with ca. 1,400 observations.
    (6) 2D Time Series Data (\textit{2D})\textsuperscript{\ref{data:ecgtd}} contains seven sets of 2-dimensional time series that are converted from trajectories of hand writings. Each set has 3 time series, each with ca. 1,000 observations.
    (7) Synthetic Data (\textit{SYN}) contains 10 univariate time series with 2,000 observations that are generated from auto-regressive processes or basis functions such as $\mathrm{sin}$ and $\mathrm{cosin}$. Then, we inject outliers into the generated time series.
    The outlier ratios (denoted as $\varphi$) of \textit{GD}, \textit{HSS}, \textit{ECG}, \textit{NAB}, \textit{S5}, \textit{2D}, and \textit{SYN} are 0.8\%, 16.7\%, 4.9\%, 9.8\%, 0.9\%, 39.2\%, and 5\%, respectively.
\rev{All datasets contains both point and collective outliers}. All datasets come with ground truth outlier labels. However, since we study unsupervised outlier detection, we do not use these labels during training. We only use them for testing, i.e., for evaluating accuracy. 
In addition, we train all methods using time series with outliers because datasets are typically not accompanied by clean time series without outliers for use in training. This setting enables us to study the robustness to outliers of different algorithms. 

\subsubsection{Baselines}
We compare with 17 existing time series outlier detection approaches: 
    (1) One-class Support Vector Machines (\texttt{OCSVM})~\cite{DBLP:journals/jmlr/ManevitzY01}, a kernel based one-class classification method;
    (2) Local Outlier Factor (\texttt{LOF})~\cite{DBLP:conf/sigmod/BreunigKNS00}, a density based outlier detection method;
    (3) Isolation Forest (\texttt{ISF})~\cite{DBLP:conf/icdm/LiuTZ08}, an unsupervised tree-based outlier detection method;
    (4) Exponential Moving Average~(\texttt{EMA})~\cite{brown1963smoothing}, a time series smoothing method using weighted moving windows, where the weights for older observations decrease exponentially.
    (5) Seasonal-Trend decomposition using Local Regression (\texttt{STL})~\cite{clevelandJOffStat1990}, a time series smoothing method that decomposes time series into trends, seasonalities, and noise;
    (6) Singular Spectrum Analysis (\texttt{SSA})~\cite{DBLP:books/daglib/NinaVA01}, a time series decomposition method using Hankelize matrix decomposition;
    (7) Matrix Profile I (\texttt{MP})~\cite{DBLP:conf/icdm/YehZUBDDSMK16}, a state-of-the-art similarity based outlier detection method;
	(10) RandNet (\texttt{RN})~\cite{DBLP:conf/sdm/ChenSAT17}, an autoencoder ensemble for outlier detection.
	(8) CNN Autoencoder (\texttt{CNNAE})~\cite{DBLP:conf/mdm/Kieu0J18}, which treats time series as images and feeds them to a 2D \texttt{CNN} autoencoder to reconstruct them;
	(9) RNN Autoencoder (\texttt{RNNAE})~\cite{DBLP:journals/corr/MalhotraRAVAS16,DBLP:conf/mdm/Kieu0J18}, which reconstructs time series with a recurrent neural network (using \texttt{LSTM} units) based autoencoder; %
	(10) BeatGAN (\texttt{BGAN})~\cite{DBLP:conf/ijcai/ZhouLHCY19}, a generative model that forms an autoencoder via adversarial learning;
	(11) Donut (\texttt{DONUT})~\cite{DBLP:conf/www/XuCZLBLLZPFCWQ18}, a variational autoencoder that reconstructs time series from stochastic latent spaces;
	(12) OmniAnomaly (\texttt{OMNI})~\cite{DBLP:conf/kdd/SuZNLSP19}, a variational recurrent autoencoder that learns stochastic latent spaces for each observation of a time series;
	(13) Transformer Autoencoder (\texttt{TAE})~\cite{TAE}, which reconstructs time series by using an attention model.
Although \texttt{LOF}, \texttt{OCSVM}, \texttt{ISF}, and \texttt{RN} were originally proposed for non-time series data, they can be applied to time series with competitive accuracy~\cite{DBLP:books/sp/Aggarwal13}. Thus, we include them in the experiments.

\subsubsection{Implementation Details}
\label{ssec:implementation_details}

All methods are implemented by using \texttt{Python 3.8}. 
Further, \texttt{PyTorch 1.1} is used for implementing all neural network based methods.
Next, \texttt{OCSVM}, \texttt{LOF}, and \texttt{ISF} are implemented using \texttt{Sklearn 1.19}, and \texttt{EMA}, \texttt{SSA}, and \texttt{MP} are implemented using \texttt{Numpy 1.15}. 
Finally, \texttt{STL} is implemented using \texttt{Statsmodels 0.12}. 
All experiments are conducted on a Linux workstation with an AMD 64-core CPU with 512 GB RAM and 2 NVIDIA Titan V GPUs.

\subsubsection{Hyperparameter Settings}
\label{ssec:hyperparameter_settings}
Since we study unsupervised outlier detection, we are unable to tune the hyperparamters using labeled data. %
To ensure fair comparisons, we consider different hyperparameter settings, and report the \emph{median} result for all methods. We do not report the best result because in unsupervised settings, we have no labeled data to enable identifying the best hyperparameters that lead to the best result. 
Specifically, we define a range for each hyperparameter. We then use random search with 200 random combinations to explore the hyperparameter space and identify a hyperparameter setting that gives the median result among all the explored hyperparameter settings. We then report this median result and consider this hyperparameter setting as the default setting.  
Next, we conduct experiments to study the sensitivity of different hyperparameters. To do so, each time, we vary a chosen hyperparameter in its range while fixing the other hyperparameters to their default settings. We proceed to provide the ranges for the hyperparameters.

For \texttt{RAE} and \texttt{RDAE}, we vary $\lambda$, $\lambda_1$, and $\lambda_2$ among $10^{-4}$, $10^{-3}$, $10^{-2}$, $10^{-1}$, and $1$; and we vary window size $B$ among $10$, $20$, $50$, $100$, $200$, and $400$. 
For all the neural network based methods, we vary the number of layers, the kernel size (\texttt{CNNAE}, \texttt{BGAN}, \texttt{RAE}, and \texttt{RDAE}), and the number of attention heads (\texttt{TAE}) among $3$, $5$, $7$, $9$, and $11$; next, we vary the number of kernels in each layer (\texttt{CNNAE}, \texttt{BGAN}, \texttt{RAE}, and \texttt{RDAE}), the number of hidden units (\texttt{RN}, \texttt{RNNAE}, \texttt{DONUT}, and \texttt{OMNI}), and the stochastic latent variable size (\texttt{DONUT} and \texttt{OMNI}) among $32$, $64$, $128$, $256$, $512$, and $1024$.
For \texttt{MP} and \texttt{EMA}, we vary the pattern size among $5$, $10$, $20$, $50$, and $100$. For \texttt{RN} and \texttt{ISF}, we vary the number of base models among $5$, $10$, $20$, $50$, $100$, and $500$. 
For \texttt{LOF}, we vary the number of neighbors among $5$, $10$, $20$, $50$, and $100$. 
For \texttt{STL}, we vary the $S$ (seasonal) and $T$ (trend) coefficients among $1$, $3$, $5$, $7$, and $9$.
For \texttt{OCSVM}, we vary the kernel degree among $3$, $5$, $7$, $9$, and $11$. 
\rev{We set $\epsilon$ to $10^{-5}$ (cf. Algorithms~1 and~2) as suggested in the literature~\cite{DBLP:journals/jacm/CandesLMW11,DBLP:journals/fa/BoydNPEJ11} that suggests $\epsilon \in [10^{-3}, 10^{-5}]$.}

\subsubsection{Evaluation Metrics} 
\label{ssec:app_evaluation_metrics}
\textbf{Accuracy and Robustness: }
Existing work typically chooses an outlier score threshold $\tau$: an observation $\mathbf{s}_i$ is considered as an outlier if $\mathcal{OS}(\mathbf{s}_i) > \tau$ and as clean data, otherwise. 
However, choosing the threshold is non-trivial and calls for domain experts or prior knowledge. Instead, following the strategy of \cite{DBLP:conf/sdm/ChenSAT17}, we employ two metrics that consider all possible thresholds---Area Under the Curve of Precision-Recall (\textit{PR})~\cite{DBLP:reference/ml/2017} and Area Under the Curve of Receiver Operating Characteristic (\textit{ROC})~\cite{DBLP:reference/ml/2017}. 
\textit{PR} and \textit{ROC} can capture the performance of an outlier detection method without the need to choose any specific threshold. 
The higher \textit{PR} and \textit{ROC}, the more robust the method is.

\noindent
\textbf{Explainability: }We use \textit{RPM}-based and \textit{SSA}-based post-hoc analysis to evaluate the explainability of different \texttt{AE}-based methods. 

\subsection{Experimental Results}%
\subsubsection{Outlier Detection Accuracy} 
Tables~\ref{tab:overall_accuracy_pr} and~\ref{tab:overall_accuracy_roc} show that \texttt{RAE} and \texttt{RDAE} achieve on average the best and second best results, respectively. \rev{On some datasets, although our methods are not the best, they do not fall behind much.}
\rev{On datasets \textit{HSS} and \textit{2D}, our proposed methods are not the best. 
This is mainly because these datasets represents trajectories, i.e., time series of locations.
More specifically, \textit{HSS} is collected from a system that contains four short conveyor belts and two rails. These belts and rails are moving when the system is working, and the time series describe the positions of the belts and rails. 
Next, \textit{2D} contains 2-dimensional time series that are converted from trajectories of hand writings. 
In such cases, distance-based and partition-based algorithms, such as \texttt{LOF} and \texttt{ISF} achieve better accuracy than the methods we propose because Euclidean space fits well with the data that represents physical locations.
We leave improvements for this kind of data to future studies.}
\rev{\texttt{RDAE} performs the best on four out of seven datasets w.r.t. \textit{PR} and \textit{ROC}.}
Further, \texttt{RDAE} performs better than \texttt{RAE} in most the cases. This indicates that the dual \texttt{AE} architecture enhances the accuracy of a single \texttt{AE}. \rev{Because \texttt{RDA} cannot capture temporal dependencies, its performance is inferior to both \texttt{RAE} and \texttt{RDAE}.}

\begin{table*}[ht]
	\fontsize{8pt}{8pt}\selectfont
	\centering
	\caption{Overall Accuracy, \textit{PR}.}
	\label{tab:overall_accuracy_pr}
	\begin{tabular}{|p{0.45cm}|P{0.55cm}P{0.55cm}P{0.50cm}P{0.55cm}P{0.55cm}P{0.55cm}P{0.50cm}P{0.50cm}P{0.65cm}P{0.65cm}P{0.55cm}P{0.65cm}P{0.55cm}P{0.55cm}P{0.55cm}|P{0.55cm}P{0.55cm}|}
		\toprule
		              & \texttt{OCSVM} & \texttt{LOF}      & \texttt{ISF} & \texttt{EMA} & \texttt{STL} & \texttt{SSA} & \texttt{MP} & \texttt{RN} & \texttt{CNNAE} & \texttt{RNNAE} & \texttt{BGAN} & \texttt{DONUT} & \texttt{OMNI}     & \texttt{TAE} & \rev{\texttt{RDA}} & \texttt{RAE}      & \texttt{RDAE}     \\ \hline
		\textit{GD}   & 0.112          & 0.079             & 0.087        & 0.081        & 0.078        & 0.134        & 0.094       & 0.112       & 0.116          & 0.128          & 0.109         & 0.097          & 0.131             & 0.088        &     \rev{0.113}    & \underline{0.141} & \textbf{0.152}    \\
		\textit{HSS}  & 0.197          & \underline{0.227} & 0.217        & 0.158        & 0.156        & 0.167        & 0.174       & 0.146       & 0.207          & \textbf{0.242} & 0.214         & 0.203          & 0.219             & 0.195    &  \rev{0.168}  & 0.197             & 0.223             \\
		\textit{ECG}  & 0.121          & 0.112             & 0.119        & 0.089        & 0.125        & 0.103        & 0.116       & 0.105       & 0.107          & 0.118          & 0.103         & 0.131          & \underline{0.144} & 0.138   &  \rev{0.115}   & 0.127             & \textbf{0.150}    \\
		\textit{NAB}  & \textbf{0.193} & 0.145             & 0.167        & 0.136        & 0.149        & 0.148        & 0.173       & 0.168       & 0.177          & 0.163          & 0.151         & 0.188          & 0.165             & 0.175    &  \rev{0.162}  & 0.159             & \underline{0.190} \\
		\textit{S5}   & 0.386          & 0.358             & 0.397        & 0.461        & 0.465        & 0.468        & 0.258       & 0.232       & 0.383          & 0.421          & 0.434         & 0.272          & 0.312             & 0.298        &    \rev{0.377}     &   \underline{0.472} & \textbf{0.481}    \\
		\rev{\textit{2D}}  & \rev{0.468} & \rev{0.476} & \rev{\underline{0.511}} & \rev{0.482} & \rev{0.435} & \rev{0.422} & \rev{0.398} & \rev{0.426} & \rev{0.401} & \rev{0.413} & \rev{0.408} & \rev{0.414} & \rev{0.427} & \rev{0.405} & \rev{0.415} & \rev{0.498} & \rev{\textbf{0.516}} \\
        \rev{\textit{SYN}} & \rev{0.102} & \rev{0.121} & \rev{0.126} & \rev{0.135} & \rev{0.129} & \rev{0.128} & \rev{0.113} & \rev{0.105} & \rev{0.116} & \rev{0.125} & \rev{0.144} & \rev{0.132} & \rev{0.118} & \rev{0.124} & \rev{0.127} & \rev{\textbf{0.161}} & \rev{\underline{0.158}} \\ \hline
		\rev{\textit{Avg.}} & \rev{0.226} & \rev{0.217} & \rev{0.232} & \rev{0.220} & \rev{0.220} & \rev{0.224} & \rev{0.189} & \rev{0.185} & \rev{0.215} & \rev{0.230} & \rev{0.223} & \rev{0.205} & \rev{0.217} & \rev{0.203} & \rev{0.211} & \rev{\underline{0.251}} & \rev{\textbf{0.267}} \\ 		\bottomrule
	\end{tabular}
\end{table*}
\begin{table*}[ht]
	\fontsize{8pt}{8pt}\selectfont
	\centering
	\caption{Overall Accuracy, \textit{ROC}.}
	\label{tab:overall_accuracy_roc}
		\begin{tabular}{|p{0.45cm}|P{0.55cm}P{0.55cm}P{0.50cm}P{0.55cm}P{0.55cm}P{0.55cm}P{0.50cm}P{0.50cm}P{0.65cm}P{0.65cm}P{0.55cm}P{0.65cm}P{0.55cm}P{0.55cm}P{0.55cm}|P{0.55cm}P{0.55cm}|}
		\toprule
		              & \texttt{OCSVM} & \texttt{LOF}      & \texttt{ISF} & \texttt{EMA} & \texttt{STL} & \texttt{SSA} & \texttt{MP} & \texttt{RN} & \texttt{CNNAE} & \texttt{RNNAE} & \texttt{BGAN} & \texttt{DONUT} & \texttt{OMNI}     & \texttt{TAE} & \rev{\texttt{RDA}} & \texttt{RAE}      & \texttt{RDAE}     \\ \hline
		\textit{GD}   & 0.582          & 0.562        & 0.651          & 0.528          & 0.552        & 0.544        & 0.545       & 0.731       & 0.641          & 0.608          & 0.673         & 0.664          & 0.658         & 0.652   &  \rev{0.613}   & \underline{0.744} & \textbf{0.763}    \\
		\textit{HSS}  & 0.523          & 0.553        & \textbf{0.618} & 0.521          & 0.537        & 0.514        & 0.528       & 0.526       & 0.560          & 0.537          & 0.549         & 0.525          & 0.557         & 0.563   &  \rev{0.550}   & 0.565             & \underline{0.581} \\
		\textit{ECG}  & 0.526          & 0.563        & 0.557          & 0.542          & 0.537        & 0.519        & 0.534       & 0.524       & 0.574          & 0.552          & 0.547         & 0.531          & 0.551         & 0.542   &  \rev{0.556}   & \underline{0.577} & \textbf{0.588}    \\
		\textit{NAB}  & 0.533          & 0.519        & 0.551          & 0.521          & 0.528        & 0.548        & 0.516       & 0.532       & 0.541          & 0.524          & 0.524         & 0.563          & 0.557         & 0.536   &  \rev{0.528}   & \underline{0.567} & \textbf{0.572}    \\
		\textit{S5}   & 0.688          & 0.733        & 0.795          & \textbf{0.835} & 0.778        & 0.767        & 0.655       & 0.575       & 0.757          & 0.753          & 0.677         & 0.678          & 0.660         & 0.635   &  \rev{0.741}   & 0.778             & \underline{0.819}             \\ 
		\rev{\textit{2D}}  & \rev{0.541} & \rev{\textbf{0.613}} & \rev{0.598} & \rev{0.528} & \rev{0.501} & \rev{0.528} & \rev{0.507} & \rev{0.523} & \rev{0.529} & \rev{0.551} & \rev{0.568} & \rev{0.573} & \rev{0.588} & \rev{0.571} & \rev{0.575} & \rev{\underline{0.611}} & \rev{0.609} \\
        \rev{\textit{SYN}} & \rev{0.549} & \rev{0.563} & \rev{0.547} & \rev{0.592} & \rev{0.598} & \rev{0.584} & \rev{0.565} & \rev{0.561} & \rev{0.603} & \rev{0.594} & \rev{0.562} & \rev{0.564} & \rev{0.569} & \rev{0.571} & \rev{0.573} & \rev{\underline{0.608}} & \rev{\textbf{0.613}} \\ \hline
		\rev{\textit{Avg.}}   & \rev{0.563} & \rev{0.587} & \rev{0.617} & \rev{0.581} & \rev{0.576} & \rev{0.572} & \rev{0.550} & \rev{0.567} & \rev{0.601} & \rev{0.588} & \rev{0.586} & \rev{0.585} & \rev{0.591} & \rev{0.581} & \rev{0.591} & \rev{\underline{0.636}} & \rev{\textbf{0.649}} \\
		\bottomrule 
	\end{tabular}
\end{table*}

\subsubsection{Performance Improvement Significance}
To determine whether the performance improvements of the proposed methods over the state-of-the-art methods are statistically significant, we conduct \textit{t-tests} to assess the significance of the proposed methods against the baselines on the average results of all datasets. The \textit{p-values} for both metrics are below 0.005. This indicates that the performance improvements over the state-of-the-art methods are statistically significant.

\subsubsection{Effect of $\lambda$} We vary $\lambda$ among $10^{-4}$, $10^{-3}$, $10^{-2}$, $10^{-1}$, and $1$. For \texttt{RDAE} we set $\lambda_1=\lambda_2$ and vary them like $\lambda$. 
We also consider \texttt{RSSA} that employs \texttt{RPCA} to replace \texttt{PCA} as used in \texttt{SSA}. 
Intuitively, a small $\lambda$ encourages more data to be kept in $\mathbf{S}$ and $\mathcal{T}_\mathbf{S}$ as outliers and forces only few observations to be considered as inliers. Instead, a large $\lambda$ encourages more observations to be kept in $\mathbf{L}$ and $\mathcal{T}_\mathbf{L}$ as inliers; thus, fewer observations are considered as outliers. Due to the space limitation, we report on effect of $\lambda$ on dataset \textit{S5} only. The results on the other datasets show similar trends. Fig.~\ref{fig:Lambda_PR_ROC} shows the results w.r.t. \textit{PR} and \textit{ROC}.
All methods achieve the best performance when $\lambda$ is between $10^{-2}$ and $10^{-1}$. When $\lambda < 10^{-2}$, some clean data is mixed with both outliers in $\mathbf{S}$ and $\mathcal{T}_\mathbf{S}$, increasing the false positives. In contrast, when $\lambda > 10^{-1}$, only a few significant outliers are kept in $\mathbf{S}$ and $\mathcal{T}_\mathbf{S}$; thus, very few outliers are detected. Although this avoids false positives, some outliers are not detected, increasing the false negatives. When $\lambda \in (10^{-2}, 10^{-1})$, the best trade-off is achieved.

\subsubsection{Effect of $B$} 
We study the effect of the window size $B$ in the frameworks \texttt{SSA} and \texttt{RDAE} that both employ lagged matrices created based on $B$. We also consider \texttt{RSSA} that employs \texttt{RPCA} to replace \texttt{PCA} as used in \texttt{SSA}. 
We vary $B$ among 10, 20, 50, 100, 200, and 400. We report findings on dataset \textit{S5} only. The results on the other datasets show similar trends. Fig.~\ref{fig:b_pr_roc} shows the results w.r.t. \textit{PR} and \textit{ROC}. 
All methods achieve the peak performance when $B=200$, which suggests that $B=200$ is the most appropriate setting for all the datasets. 
This observation is consistent with the recommendation by Khan and Poskitt~\cite{Scholar:jour/monash/Khan11} \rev{that $B$ is set to $(\mathrm{ln} \; C)^\psi$, where $\psi\in(1.5, 3.0)$ and $C$ is the length of a time series (see Section II). In \textit{S5}, $C$ is around 1,400.}

\subsubsection{Ablation Study} 
\begin{figure*}[t]
\centering
\begin{minipage}{0.33\textwidth}
    \centering
	\begin{subfigure}[b]{0.49\linewidth}
		\includegraphics[clip, trim=0.3cm 0.4cm 0.2cm 0.5cm, width=1.05\linewidth]{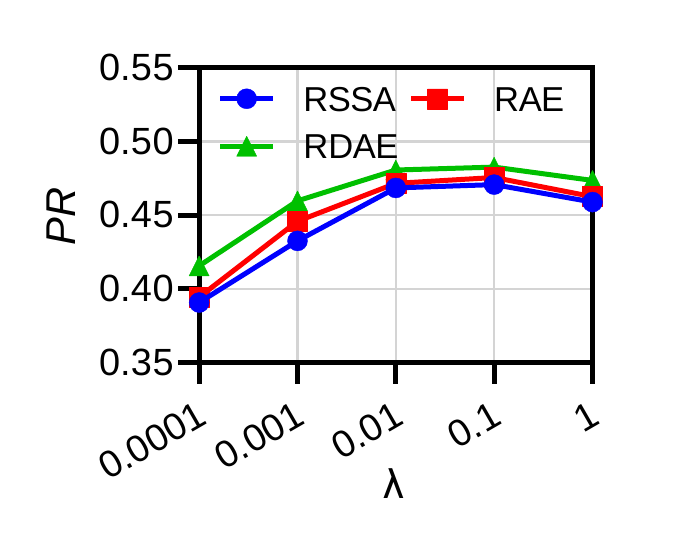}
		\caption{\textit{PR}.}
		\label{fig:lambda_pr_yahoo}
	\end{subfigure}
	\begin{subfigure}[b]{0.49\linewidth}
		\includegraphics[clip, trim=0.3cm 0.4cm 0.2cm 0.5cm, width=1.05\linewidth]{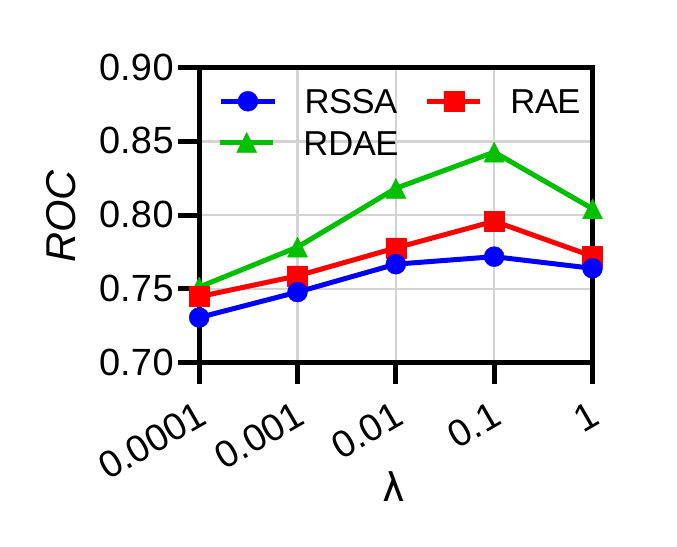}
		\caption{\textit{ROC}.}
		\label{fig:lambda_roc_yahoo}
	\end{subfigure}
	\caption{Effect of $\lambda$.}
	\label{fig:Lambda_PR_ROC}
\end{minipage}%
\begin{minipage}{0.33\textwidth}
    \centering
	\begin{subfigure}[b]{0.49\linewidth}
		\includegraphics[clip, trim=0.3cm 0.4cm 0.2cm 0.5cm, width=1.1\linewidth]{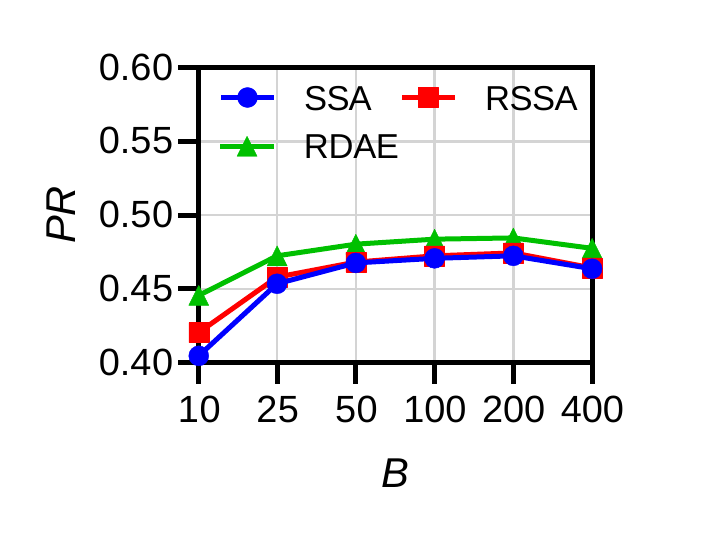}
		\caption{\textit{PR}.}
		\label{fig:b_pr_yahoo}
	\end{subfigure}
	\begin{subfigure}[b]{0.49\linewidth}
		\includegraphics[clip, trim=0.3cm 0.4cm 0.2cm 0.5cm, width=1.1\linewidth]{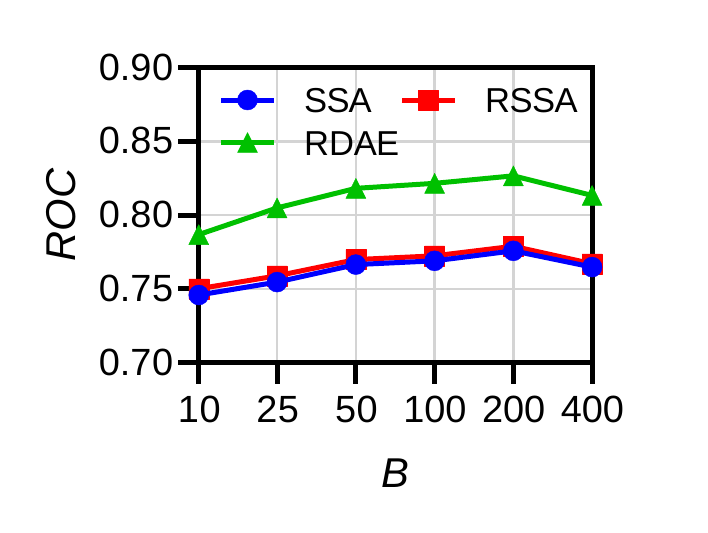}
		\caption{\textit{ROC}.}
		\label{fig:b_roc_yahoo}
	\end{subfigure}
	\caption{Effect of $B$.}
	\label{fig:b_pr_roc}
\end{minipage}%
\begin{minipage}{0.33\textwidth}
    \centering
	\begin{subfigure}[b]{0.49\linewidth}
		\includegraphics[clip, trim=0.3cm 0.4cm 0.2cm 0.5cm, width=1.05\linewidth]{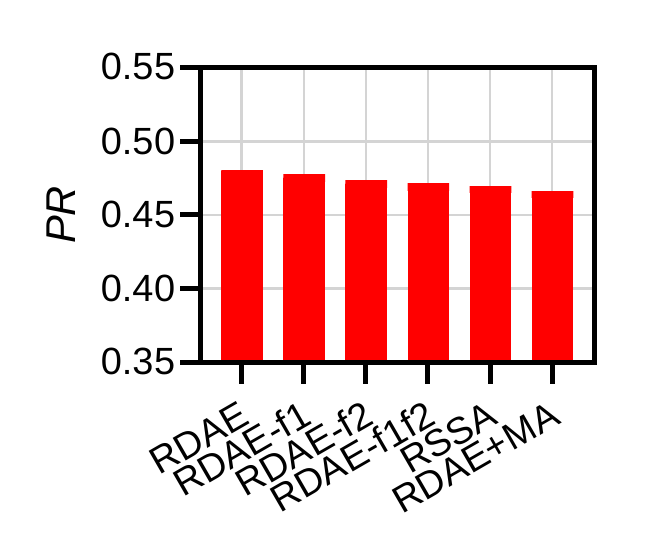}
		\caption{\textit{PR}.}
		\label{fig:ab_pr_yahoo}
	\end{subfigure}
	\begin{subfigure}[b]{0.49\linewidth}
		\includegraphics[clip, trim=0.3cm 0.4cm 0.2cm 0.5cm, width=1.05\linewidth]{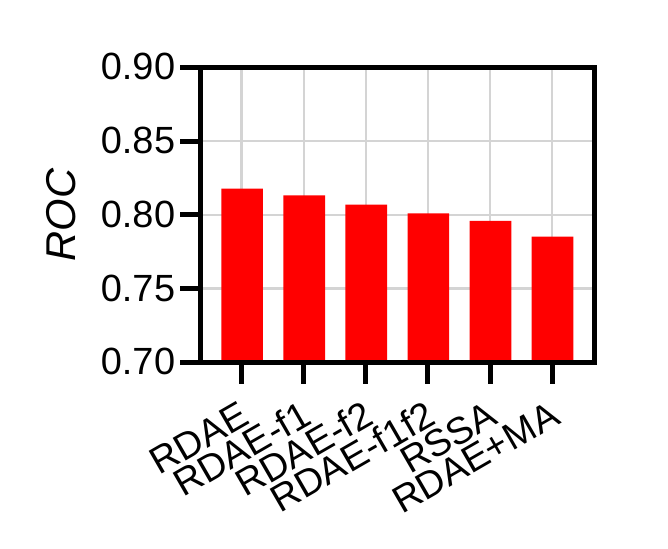}
		\caption{\textit{ROC}.}
		\label{fig:ab_roc_yahoo}
	\end{subfigure}
	\caption{Ablation Study.}
	\label{fig:ablation}
\end{minipage}%
\end{figure*}

We study the effect of each component in \texttt{RDAE} by removing $f_{1_{\theta_1}}(\cdot)$ (denoted as \texttt{RDAE-f1}), removing $f_{2_{\theta_2}}(\cdot)$ (denoted as \texttt{RDAE-f2}), and removing both (denoted as \texttt{RDAE-f1f2}). \rev{\texttt{RDAE-f1f2} can be viewed as \texttt{RDA}~\cite{DBLP:conf/kdd/ZhouP17}.} We also consider \texttt{RSSA} that employs \texttt{RPCA} to replace \texttt{PCA} as used in \texttt{SSA}.
We report findings on dataset \textit{S5} only. The results on the other datasets show similar trends. Fig.~\ref{fig:ablation} shows the results w.r.t. \textit{PR} and \textit{ROC}. 
\rev{\texttt{RDAE} outperforms \texttt{RAE} that only uses the time series view, suggesting that multi-view (i.e., both time series and matrix representation) is superior to single-view (i.e., only time series representation). \texttt{RDAE} outperforms \texttt{RDAE-f1f2} that only uses the (single-view) lagged matrix.}
\texttt{RDAE-f1} and \texttt{RDAE-f2} outperform \texttt{RDAE-f1f2} w.r.t. \textit{PR} and \textit{ROC}. 
Further, \texttt{RDAE-f1} outperforms \texttt{RDAE-f2} w.r.t. \textit{PR} and \textit{ROC}, which suggests that the outer \texttt{AE} $f_{2_{\theta_2}}(\cdot)$ is more important than $f_{1_{\theta_1}}(\cdot)$ because that is only a non-linear transformation that offers input for the inner \texttt{AE}, while $f_{2_{\theta_2}}(\cdot)$ performs the reconstruction of the time series view. %
\texttt{RDAE} achieves the best performance, which suggests that the two \texttt{AE}s together with the additional nonlinear transformation offered by $f_{1_{\theta_1}}(\cdot)$ yields the most powerful framework. 
\rev{Further, we consider \texttt{RDAE+MA}, where we use a simple moving average to smooth the input time series $\mathcal{T}$. 
The smoothed time series is then transformed to the smoothed lagged matrix $\hat{\mathbf{M}}$. The results show that \texttt{RDAE+MA} is worse than \texttt{RDAE}, suggesting that our proposed non-linear transformation is better than moving average smoothing.}

\subsubsection{Robustness}
We study the robustness of the proposed methods by removing the robustness considerations of \texttt{RAE} and \texttt{RDAE} to create two non-robust counterparts, denoted as \texttt{N-RAE} and \texttt{N-RDAE}, respectively. Specifically, for \texttt{N-RAE}, we employ an 1D \texttt{CNN}-based \texttt{AE} to reconstruct time series $\mathcal{T}$. The reconstructed time series is considered as the clean time series $\mathcal{T}_\mathbf{L}$. Next, the outlier scores are computed by differencing $\mathcal{T}$ and $\mathcal{T}_\mathbf{L}$. For \texttt{N-RDAE}, we employ a 2D \texttt{CNN}-based \texttt{AE} as the inner \texttt{AE} to reconstruct the lagged matrix $\mathbf{\hat{M}}$. The reconstruction is considered as the clean lagged matrix $\mathbf{L}$. Then, we employ an 1D \texttt{CNN}-based \texttt{AE} as the outer \texttt{AE} to reconstruct the time series, which is obtained from $\mathbf{L}$. The reconstructed time series is considered as the clean time series $\mathcal{T}_\mathbf{L}$. As before, the outlier scores are computed by differencing $\mathcal{T}$ and $\mathcal{T}_\mathbf{L}$. Fig.~\ref{fig:robustness} shows the experimental results of \texttt{RAE} vs. \texttt{N-RAE} and \texttt{RDAE} vs. \texttt{N-RDAE} w.r.t. \textit{PR} and \textit{ROC}. We report results on dataset \textit{S5} only. The results on the other datasets show similar trends. The results show that \texttt{RAE} outperforms \texttt{N-RAE} and that \texttt{RDAE} outperforms \texttt{N-RDAE}, justifying the robustness design choices used in \texttt{RAE} and \texttt{RDAE}. \rev{Further, we observe that 
many time series in experimental datasets contain only few outliers but these outliers can affect the non-robust \texttt{AE}s, thus making \texttt{N-RAE} and \texttt{N-RDAE} work poorly}. 

\subsubsection{Effect of Different Architectures}
\label{ssec:effect_fc}
\begin{figure*}[h]
\centering
\begin{minipage}[b]{0.33\textwidth}
    \begin{subfigure}[b]{0.49\linewidth}
		\includegraphics[clip, trim=0.3cm 0.4cm 0.2cm 0.5cm, width=1.0\linewidth]{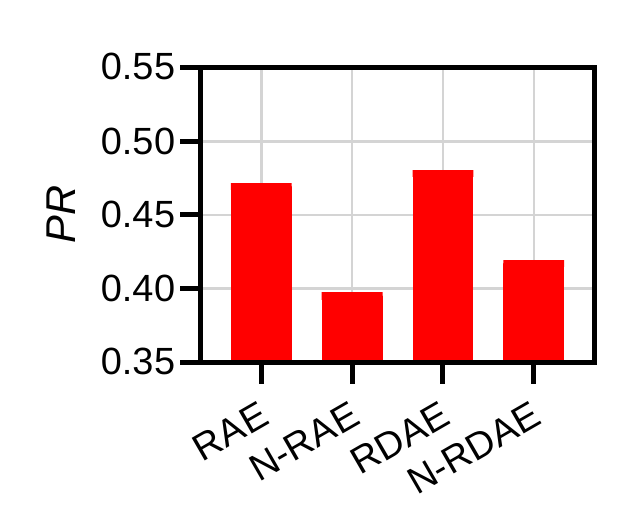}
		\caption{\textit{PR}.}
		\label{subfig:robustness_pr_yahoo}
	\end{subfigure}
	\begin{subfigure}[b]{0.49\linewidth}
		\includegraphics[clip, trim=0.3cm 0.4cm 0.2cm 0.5cm, width=1.0\linewidth]{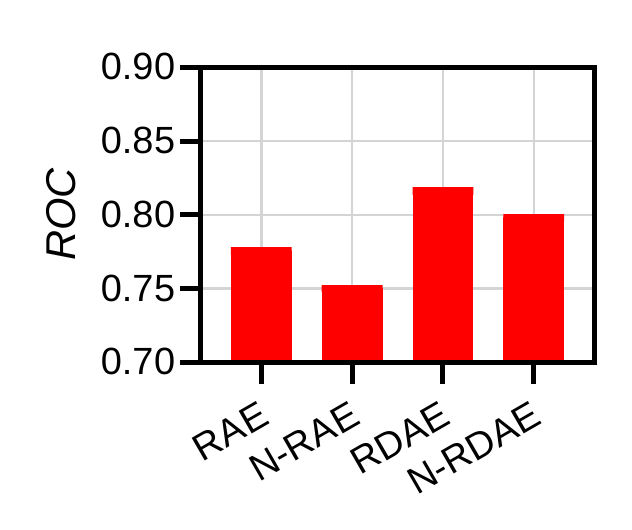}
		\caption{\textit{ROC}.}
		\label{subfig:robustness_roc_yahoo}
	\end{subfigure}
	\caption{Robustness.}
	\label{fig:robustness}
\end{minipage}%
\begin{minipage}[b]{0.5\textwidth}
    \begin{subfigure}[b]{0.33\linewidth}
		\includegraphics[clip, trim=0.3cm 0.4cm 0.2cm 0.5cm, width=1.0\linewidth]{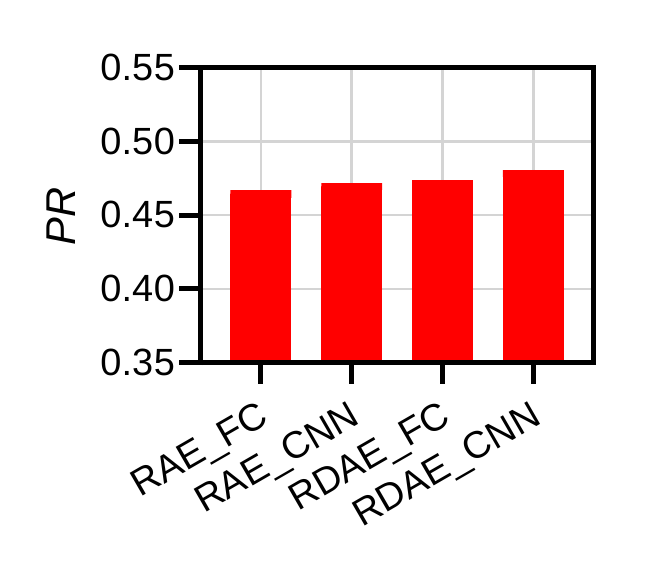}
		\caption{\textit{PR}.}
		\label{subfig:different_architecture_pr_yahoo}
	\end{subfigure}
	\begin{subfigure}[b]{0.32\linewidth}
		\includegraphics[clip, trim=0.3cm 0.4cm 0.2cm 0.5cm, width=1.0\linewidth]{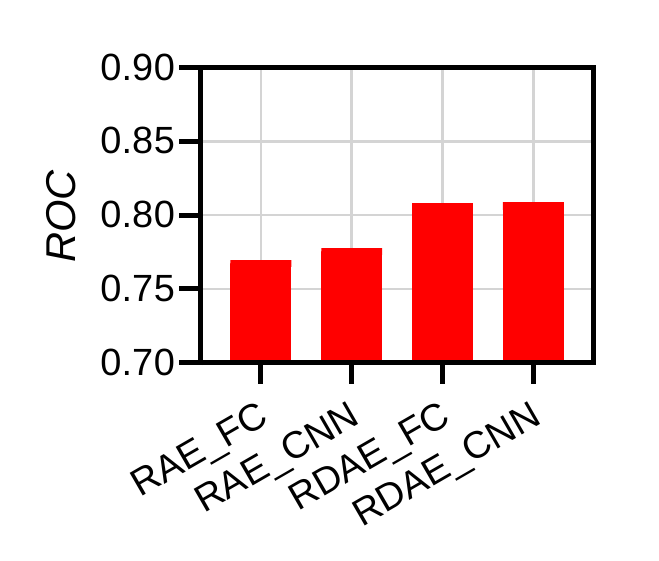}
		\caption{\textit{ROC}.}
		\label{subfig:different_architecture_roc_yahoo}
	\end{subfigure}
	\begin{subfigure}[b]{0.32\linewidth}
		\includegraphics[clip, trim=0.3cm 0.4cm 0.2cm 0.5cm, width=0.94\linewidth]{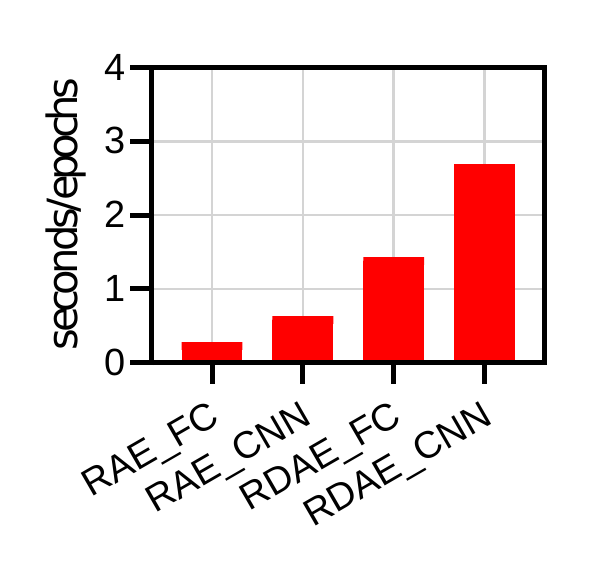}
		\caption{\textit{Runtime}.}
		\label{subfig:different_architecture_runtime_yahoo}
	\end{subfigure}
	\caption{Effect of the Different Architectures.}
	\label{fig:different_architecture}
\end{minipage}%
\end{figure*}

It is worth noting that \texttt{RAE} and \texttt{RDAE} are generic architectures rather than specific models. We thus study the effect of using fully-connected layers instead of \texttt{CNN} layers in \texttt{RAE} and \texttt{RDAE}. We implement \texttt{RAE} and \texttt{RDAE} by using fully-connected layers instead of \texttt{CNN} layers.
Fig.~\ref{fig:different_architecture} shows the experimental results of \texttt{RAE} using the fully-connected layers (denoted as \texttt{RAE\_FC}), \texttt{RAE} using the \texttt{CNN} layers (denoted as \texttt{RAE\_CNN}), \texttt{RDAE} using the fully-connected layers (denoted as \texttt{RDAE\_FC}), and \texttt{RDAE} using the \texttt{CNN} layers (denoted as \texttt{RDAE\_CNN}) w.r.t. \textit{PR}, \textit{ROC}, and training \textit{runtime}. We report results on dataset \textit{S5} only. The results on the other datasets show similar trends.
We see that the runtimes of our proposals can be reduced by using fully-connected layers while achieving competitive results. 
This shows that our generic design offers flexibility regarding runtime vs. accuracy.
When training time is critical, \texttt{RAE\_FC} is a desirable choice.
We believe that our frameworks can run much faster by employing pre-trained frameworks~\cite{DBLP:conf/cvpr/JainL11} and fine-tuning
techniques~\cite{DBLP:conf/iclr/LiCYLRBS20} instead of training every framework from scratch.

\subsubsection{Effect of the Number of Hidden Layers} 
\label{ssec:app_number_layers}
We study the effect of the number of hidden \texttt{CNN} layers in \texttt{RAE} and \texttt{RDAE}.
Specifically, we vary the number of hidden layers among 3, 5, 7, 9, and 11. We report findings on dataset \textit{S5} only. The results on the other datasets exhibit similar trends. Fig.~\ref{fig:n_pr_roc} shows the results w.r.t. \textit{PR} and \textit{ROC}. 
The results show that the frameworks with more hidden layers achieve slightly better accuracy. This indicates that the number of hidden layers is insensitive to the accuracy. This suggests that while we cannot tune this hyperparameter, randomly chosen this hyperparameter yields good accuracy.

\subsubsection{Effect of the Number of Kernels} 
\label{ssec:app_number_kernels}
We study the effect of the number of \texttt{CNN} kernels in each layer in \texttt{RAE} and \texttt{RDAE}.
Specifically, we vary the number of kernels in each \texttt{CNN} layer among $32$, $64$, $128$, $256$, $512$, and $1024$. Besides, we ensure that the bottleneck layer has the least number of kernels compare to the other layers. We report findings on dataset \textit{S5} only. The results on the other datasets exhibit similar trends. Fig.~\ref{fig:nk_pr_roc} shows the results w.r.t. \textit{PR} and \textit{ROC}. 
The results show that the frameworks with more kernels in each layer achieve slightly better accuracy. This indicates that the number of kernel in each layer is insensitive to the accuracy. This suggests that while we cannot tune this hyperparameter, randomly chosen this hyperparameter yields good accuracy. 

\subsubsection{Effect of the Kernel Size} 
\label{ssec:app_kernel_sizes}
\begin{figure*}[h]
\centering
\begin{minipage}{0.33\textwidth}
	\begin{subfigure}[b]{0.49\linewidth}
		\includegraphics[clip, trim=0.3cm 0.4cm 0.2cm 0.5cm, width=1.1\linewidth]{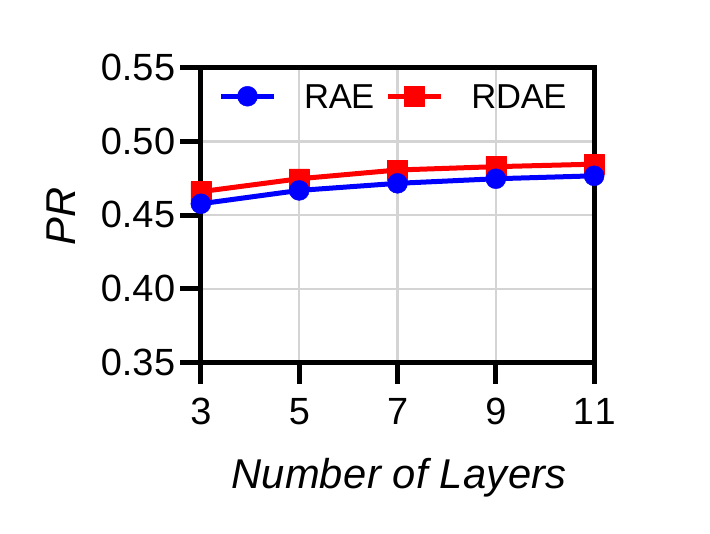}
		\caption{\textit{PR}.}
		\label{subfig:n_pr_yahoo}
	\end{subfigure}
	\begin{subfigure}[b]{0.49\linewidth}
		\includegraphics[clip, trim=0.3cm 0.4cm 0.2cm 0.5cm, width=1.1\linewidth]{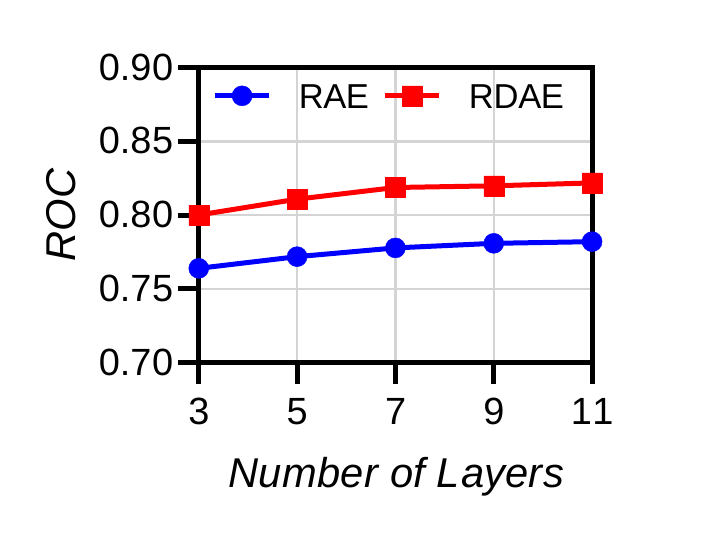}
		\caption{\textit{ROC}.}
		\label{subfig:n_roc_yahoo}
	\end{subfigure}
	\caption{Effect of the Number of Layers.}
	\label{fig:n_pr_roc}
\end{minipage}%
\begin{minipage}{0.33\textwidth}
    \centering
	\begin{subfigure}[b]{0.49\linewidth}
		\includegraphics[clip, trim=0.3cm 0.4cm 0.2cm 0.5cm, width=1.1\linewidth]{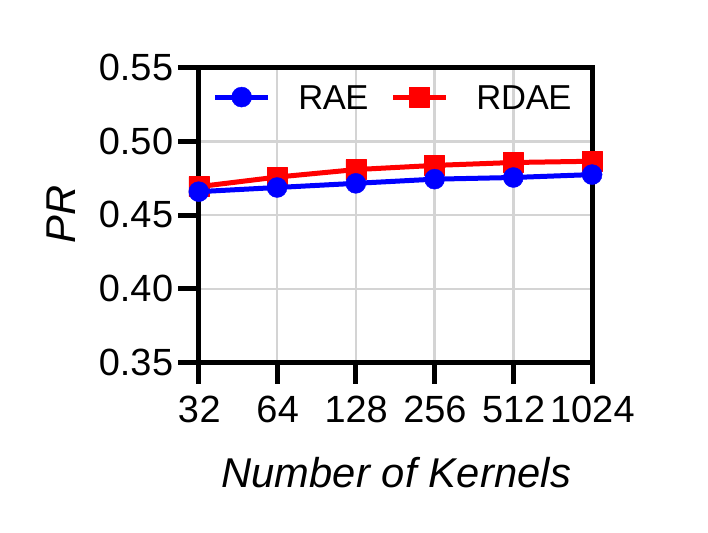}
		\caption{\textit{PR}.}
		\label{subfig:nk_pr_yahoo}
	\end{subfigure}
	\begin{subfigure}[b]{0.49\linewidth}
		\includegraphics[clip, trim=0.3cm 0.4cm 0.2cm 0.5cm, width=1.1\linewidth]{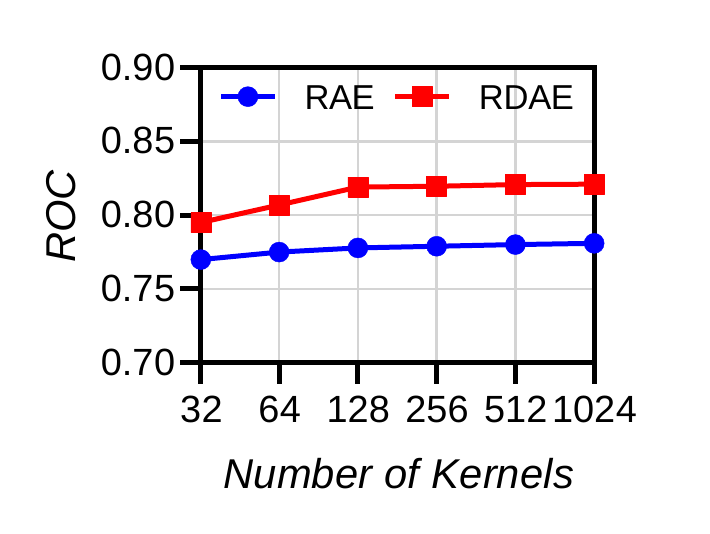}
		\caption{\textit{ROC}.}
		\label{subfig:nk_roc_yahoo}
	\end{subfigure}
	\caption{Effect of the Number of Kernels.}
	\label{fig:nk_pr_roc}
\end{minipage}%
\begin{minipage}{0.33\textwidth}
	\begin{subfigure}[b]{0.49\linewidth}
		\includegraphics[clip, trim=0.3cm 0.4cm 0.2cm 0.5cm, width=1.1\linewidth]{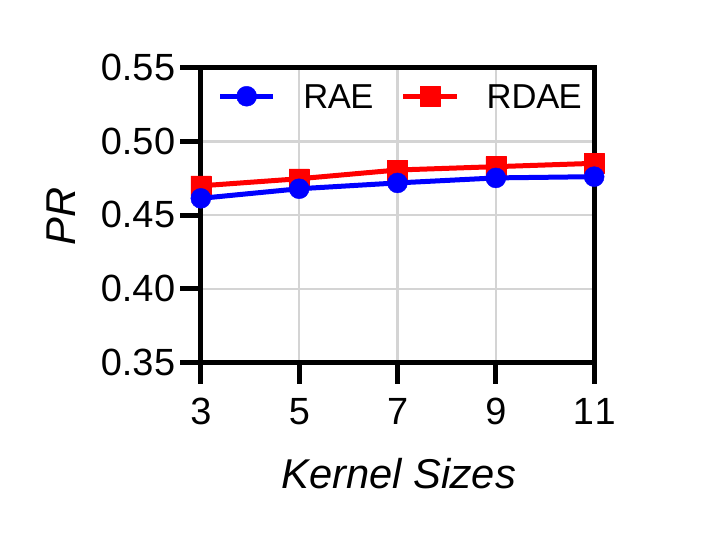}
		\caption{\textit{PR}.}
		\label{subfig:k_pr_yahoo}
	\end{subfigure}
	\begin{subfigure}[b]{0.49\linewidth}
		\includegraphics[clip, trim=0.3cm 0.4cm 0.2cm 0.5cm, width=1.1\linewidth]{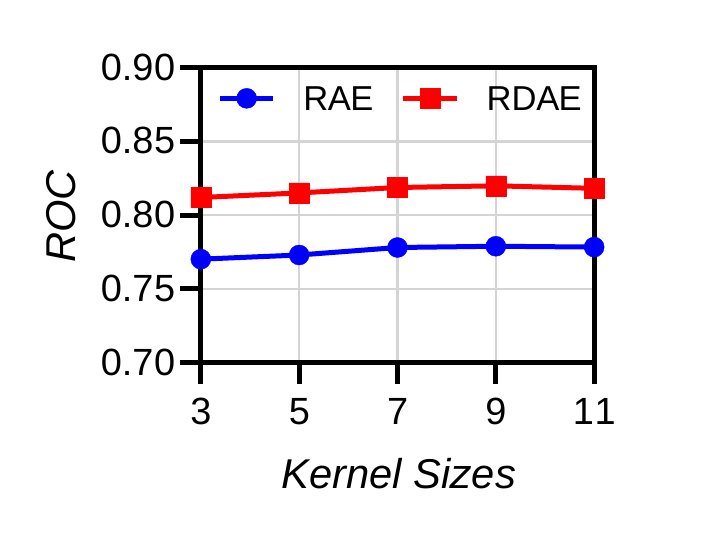}
		\caption{\textit{ROC}.}
		\label{subfig:k_roc_yahoo}
	\end{subfigure}
	\caption{Effect of the Kernel Size.}
	\label{fig:k_pr_roc}
\end{minipage}%
\end{figure*}

We study the effect of the kernel size in the \texttt{CNN} layers in \texttt{RAE} and \texttt{RDAE}. 
Specifically, we vary the kernel size among 3, 5, 7, 9, and 11. We report findings on dataset \textit{S5} only. The results on the other datasets exhibit similar trends. Fig.~\ref{fig:k_pr_roc} shows the results w.r.t. \textit{PR} and \textit{ROC}. 
The results show that the frameworks with larger kernel achieve slightly better accuracy. This indicates that the kernel size is insensitive to the accuracy. This suggests that while we cannot tune this hyperparameter, randomly chosen this hyperparameter yields good accuracy.

\begin{figure*}[h]
\centering
\begin{minipage}[b]{0.33\textwidth}
    \centering
	\begin{subfigure}[b]{0.49\linewidth}
		\includegraphics[clip, trim=0.3cm 0.4cm 0.2cm 0.2cm, width=1.0\linewidth]{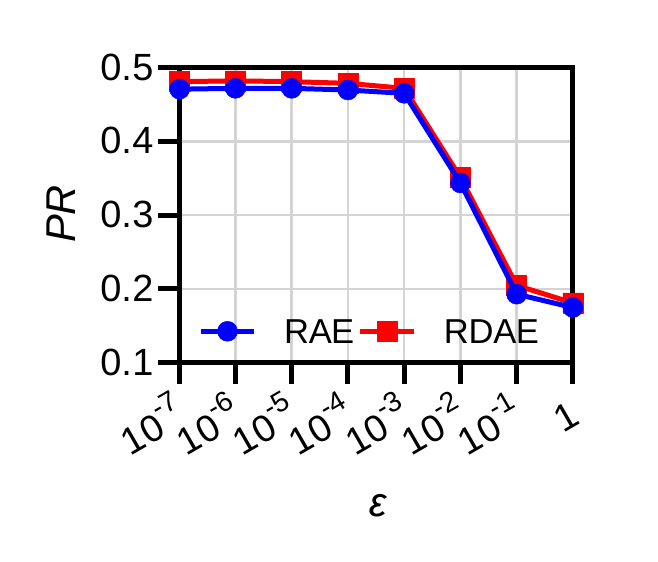}
		\caption{\rev{\textit{PR}.}}
		\label{subfig:epsilon_pr_yahoo}
	\end{subfigure}
	\begin{subfigure}[b]{0.49\linewidth}
		\includegraphics[clip, trim=0.3cm 0.4cm 0.2cm 0.2cm, width=1.0\linewidth]{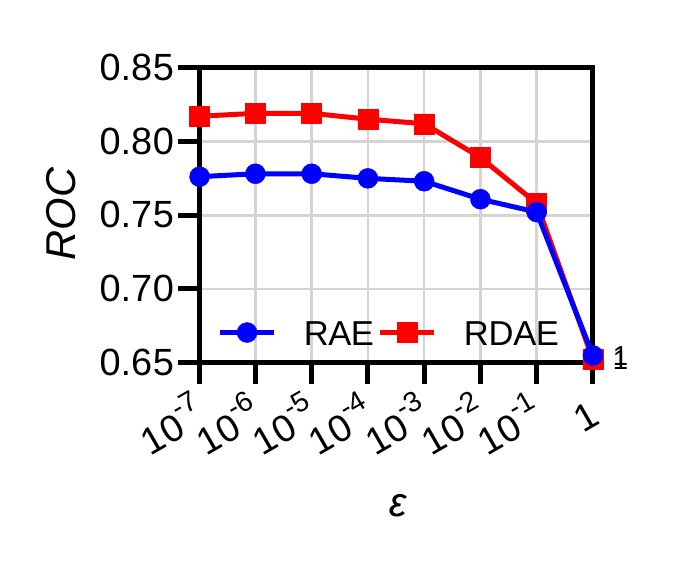}
		\caption{\rev{\textit{ROC}.}}
		\label{subfig:epsilon_roc_yahoo}
	\end{subfigure}
	\caption{\rev{Effect of $\epsilon$.}}
	\label{fig:epsilon}
\end{minipage}%
\begin{minipage}[b]{0.33\textwidth}
    \centering
        \begin{subfigure}[b]{0.49\linewidth}
    		\includegraphics[clip, trim=0.3cm 0.4cm 0.2cm 0.2cm, width=1.0\linewidth]{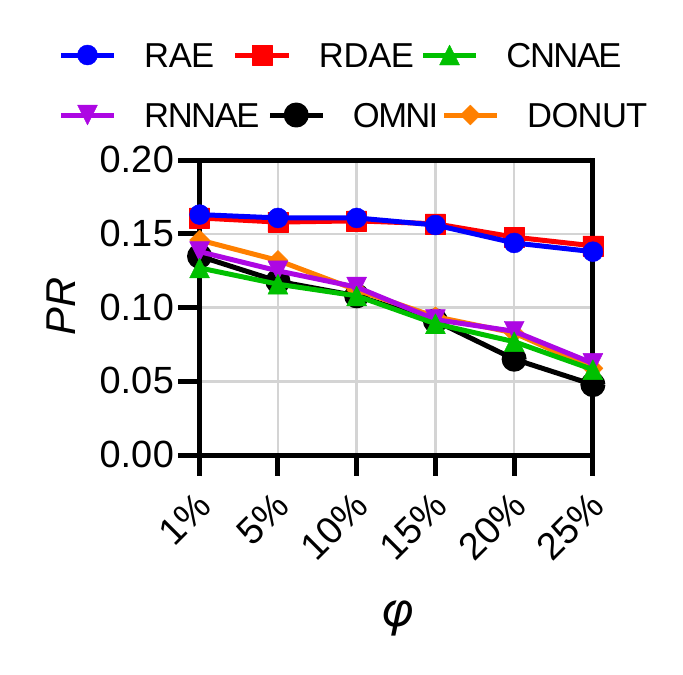}
    		\caption{\rev{\textit{PR}.}}
    		\label{subfig:varphi_pr_yahoo}
    	\end{subfigure}
    	\begin{subfigure}[b]{0.49\linewidth}
    		\includegraphics[clip, trim=0.3cm 0.4cm 0.2cm 0.2cm, width=1.0\linewidth]{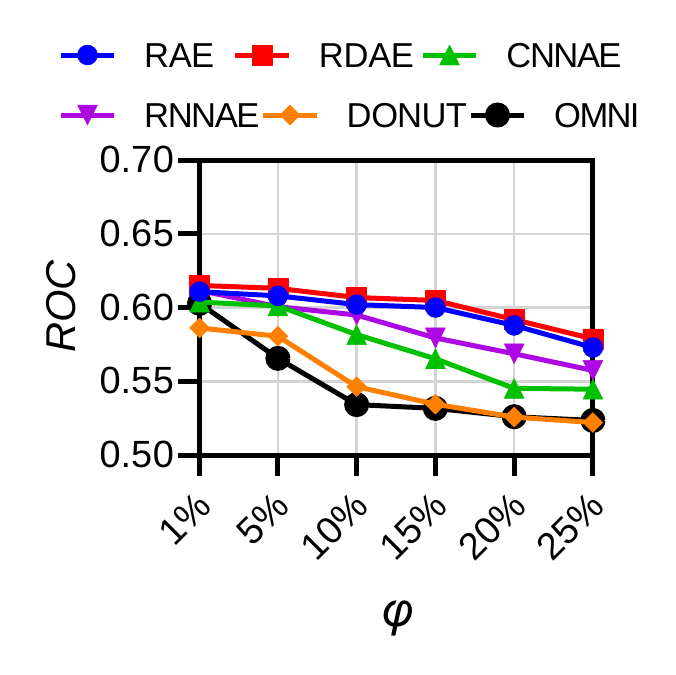}
    		\caption{\rev{\textit{ROC}.}}
    		\label{subfig:varphi_roc_yahoo}
    	\end{subfigure}
	\caption{\rev{Effect of $\varphi$.}}
	\label{fig:varphi}
\end{minipage}%
\end{figure*}

\begin{figure*}[h!]
\centering
\begin{minipage}[b]{0.33\textwidth}
    \centering
	\begin{subfigure}[b]{0.49\linewidth}
		\includegraphics[clip, trim=0.3cm 0.4cm 0.2cm 0.2cm, width=1.0\linewidth]{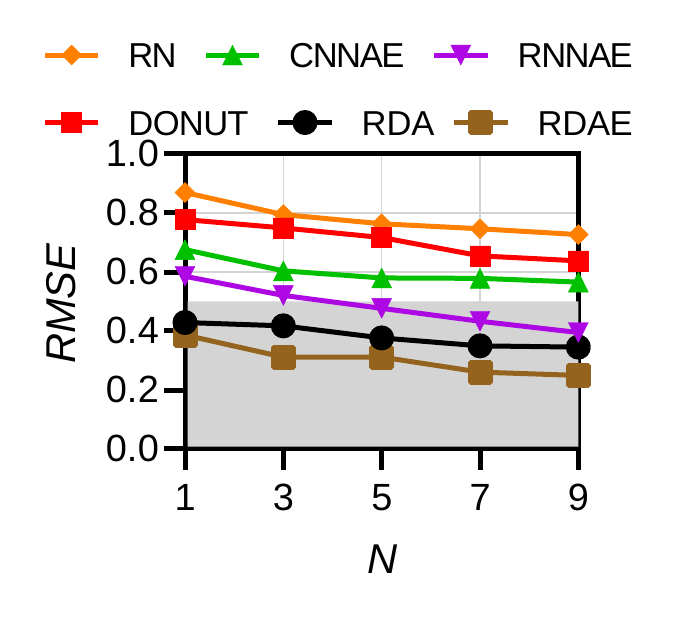}
		\caption{\textit{PHE-PRM}.}
		\label{subfig:explainability_p_yahoo}
	\end{subfigure}
	\begin{subfigure}[b]{0.49\linewidth}
		\includegraphics[clip, trim=0.3cm 0.4cm 0.2cm 0.2cm, width=1.0\linewidth]{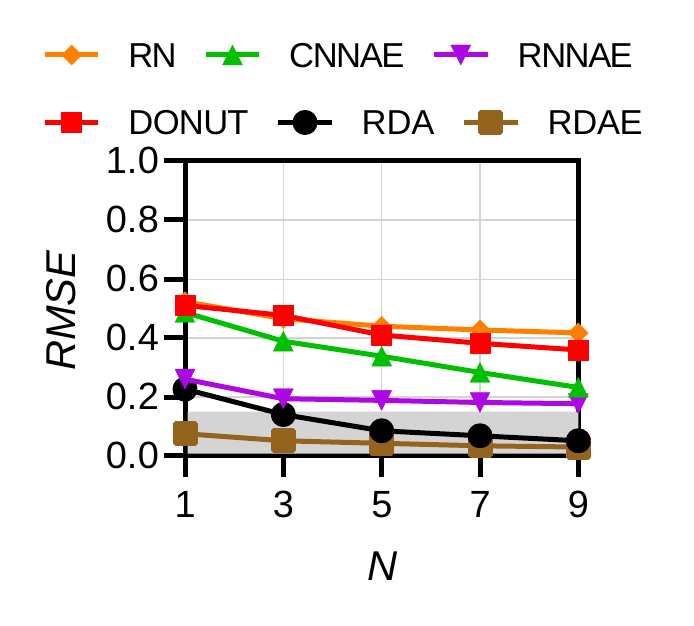}
		\caption{\textit{PHE-SSA}.}
		\label{subfig:explainability_s_yahoo}
	\end{subfigure}
	\caption{Post-hoc Explainability.}
	\label{fig:explainability_evaluation}
\end{minipage}%
\begin{minipage}[b]{0.5\textwidth}
    \centering
    \begin{subfigure}[t]{0.325\linewidth}
		\includegraphics[clip, trim=0.4cm 0.4cm 0.5cm 0.4cm, width=1.0\linewidth]{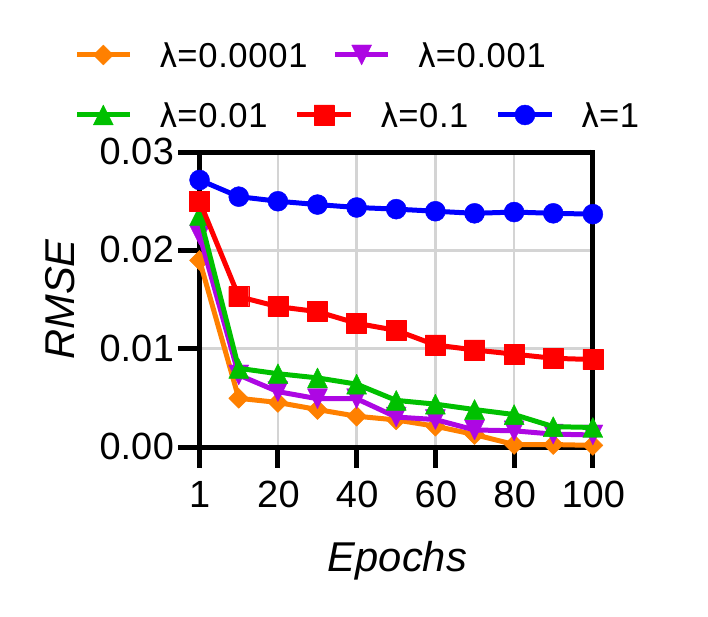}
		\caption{$\lambda$, \texttt{RAE}.}
		\label{subfig:convergence_rae_yahoo}
	\end{subfigure}
    \begin{subfigure}[t]{0.325\linewidth}
		\includegraphics[clip, trim=0.4cm 0.4cm 0.5cm 0.4cm, width=1.0\linewidth]{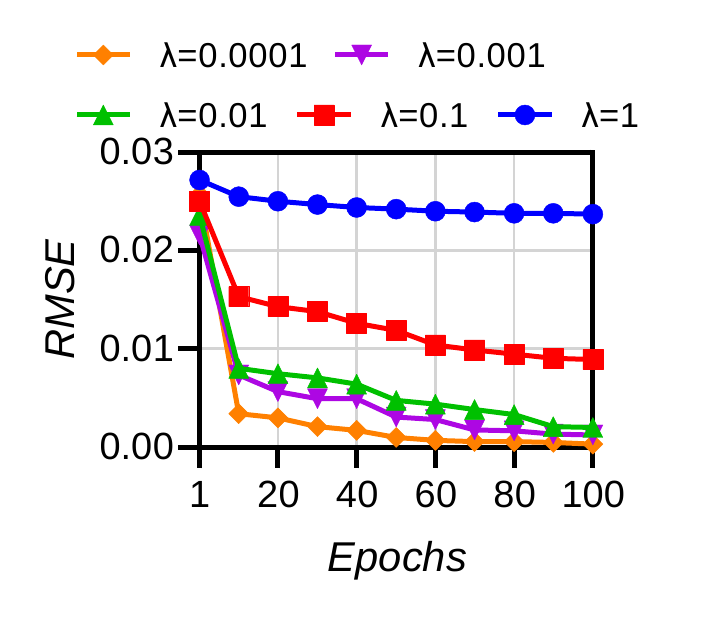}
		\caption{$\lambda$, \texttt{RDAE}.}
		\label{subfig:convergence_rdae_yahoo_a}
	\end{subfigure}
	\begin{subfigure}[t]{0.325\linewidth}
		\includegraphics[clip, trim=0.4cm 0.4cm 0.5cm 0.4cm, width=1.0\linewidth]{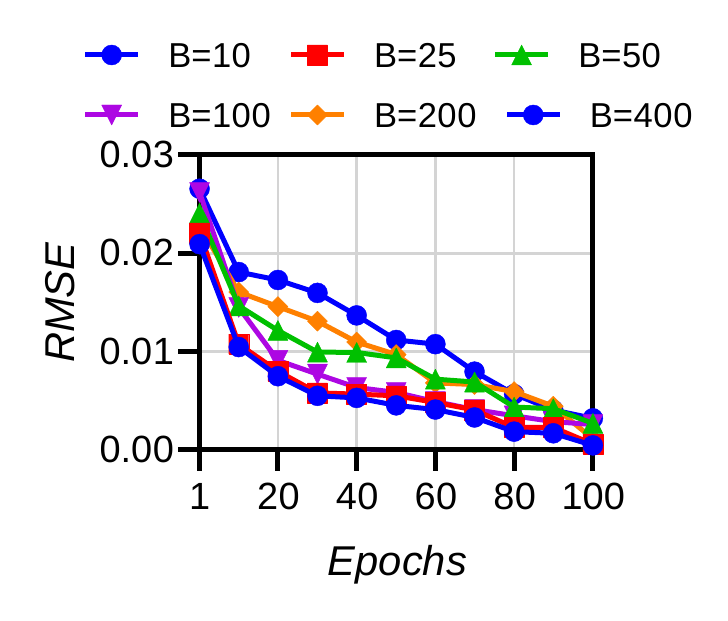}
		\caption{$B$, \texttt{RDAE}.}
		\label{subfig:convergence_rdae_yahoo_b}
	\end{subfigure}
	\caption{Effect of $\lambda$ and $B$, Convergence Analysis.}
	\label{fig:convergence_analysis_rae_rdae}
\end{minipage}%
\begin{minipage}[b]{.165\textwidth}
  \centering
		\includegraphics[clip, trim=0.3cm 0.4cm 0.2cm 0.2cm, width=1.0\linewidth]{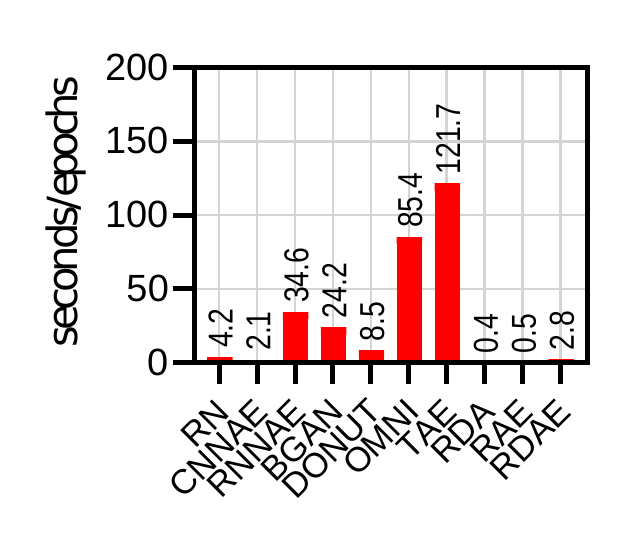}
		\label{fig:runtime_S5}
	\caption{\rev{Runtimes}.}
	\label{fig:runtime_comparision}
\end{minipage}
\end{figure*}

\subsubsection{\rev{Effect of $\epsilon$}}
\rev{We study the effect of $\epsilon$ in \texttt{RAE} and \texttt{RDAE}. Specifically, we vary $\epsilon$ among $10^{-7}$, $10^{-6}$, $10^{-5}$, $10^{-4}$, $10^{-3}$, $10^{-2}$, $10^{-1}$, and $1$. We report findings on dataset \textit{S5} only. The results on the other datasets exhibit similar trends. Fig.~\ref{fig:epsilon} shows the results w.r.t. \textit{PR} and \textit{ROC}. We observe that when $\epsilon \in [10^{-3}, 10^{-5}]$, the accuracy only changes insignificantly. When $\epsilon > 10^{-3}$, the accuracy drops because the training has not converged yet. When $\epsilon < 10^{-5}$, the accuracy does not change much but the training takes a very long time. Thus, our choice $\epsilon=10^{-5}$ is consistent with the literature~\cite{DBLP:journals/jacm/CandesLMW11,DBLP:journals/fa/BoydNPEJ11}.
}

\subsubsection{\rev{Effect of Outlier Ratios}}
\rev{We study the effect of the outlier ratios $\varphi$ of a dataset. Specifically, we vary $\varphi$ on \textit{SYN} among 1\%, 5\%, 10\%, 15\%, 20\%, and 25\%. 
We include \texttt{CNNAE}, \texttt{RNNAE}, \texttt{DONUT}, and \texttt{OMNI} as baselines due to their competitive accuracy (cf. Tables~\ref{tab:overall_accuracy_pr} and \ref{tab:overall_accuracy_roc}). 
Fig.~\ref{fig:varphi} shows the results w.r.t. \textit{PR} and \textit{ROC}. 
We see that \texttt{RAE} and \texttt{RDAE} can maintain their accuracy while the accuracy of the baselines drop quickly. This suggests that our proposed methods offer added robustness to contaminated data.}

\subsubsection{Explainability} We conduct a post-hoc explainability analysis to gain insight into the explainability of the \texttt{AE}-based methods: \texttt{CNNAE}, \texttt{RNNAE}, \texttt{RN}, \texttt{DONUT}, \texttt{RAE}, and \texttt{RDAE}. For \texttt{RAE} and \texttt{RDAE}, we consider $\mathcal{T}_\mathbf{L}$ as the clean time series. For \texttt{CNNAE}, \texttt{RNNAE}, and \texttt{DONUT}, we consider the reconstructed time series as the clean time series. For \texttt{RN}, we use the average of all reconstructed time series from the ensemble as the clean time series. Given a clean time series, we employ \textit{PHE-PRM} and \textit{PHE-SSA} (cf. Section~\ref{sec:explainability}) to quantify the explainability of the frameworks when varying  $N$ among 1, 3, 5, 7, and 9. 
\rev{Recall the \textit{RMSE} threshold $\gamma$ in Eqs.~\ref{fn:explain_prm} and~\ref{fn:explain_ssa}. We aim to identify the smallest $N$ such that the \textit{RMSE} is at most $\gamma$. 
In other words, with a specific threshold $\gamma$, the model with the smaller $N$ that can produce an \textit{RMSE} that is at most $\gamma$ is more explainable.
Fig.~\ref{fig:explainability_evaluation} shows the \textit{PHE-PRM} and \textit{PHE-SSA} results on the \textit{S5} dataset. No matter which $\gamma$ we choose, \texttt{RAE} and \texttt{RDAE} have the smallest (and thus the best) explainability scores.} 
For example, for \textit{PHE-PRM} and with $\gamma = 0.5$, \texttt{RAE} and \texttt{RDAE} both have explainability score 1, meaning that a linear function can approximate the clean time series within $\mathit{RMSE}$ 0.5. In contrast, \texttt{RNNAE} needs to use a polynomial function of degree 5. Further, \texttt{CNNAE}, \texttt{DONUT}, and \texttt{RN} fail to achieve an $\mathit{RMSE}$ below 0.5 when using up to degree 9 polynomial functions.
The results show that the proposed methods, which take into account robustness, excel at learning fundamental patterns of time series and thus are easier to explain.  
This indicates that \texttt{RAE} and \texttt{RDAE} are the most explainable and are able to learn fundamental patters of time series. The observation also holds when using \textit{PHE-SSA}. When choosing $\gamma = 0.15$, \texttt{RDAE} has explainability score 1, indicating that \texttt{RDAE} is able to produce a clean time series that can be explained by a linear trend (i.e., the most important \texttt{SSA} component). \texttt{RAE} is the second best with an explainability score of 3. The other methods are not explainable by up to 9 \texttt{SSA} components when $\gamma = 0.15$. 

We can also consider explainability from the different perspective, where we fix $N$. 
We observe that \texttt{RDAE} and \texttt{RAE} always have the lowest $\mathit{RMSE}$ values. 
This suggest that when using a specific post-hoc analysis model (i.e., a fixed $N$) to fit the clean time series, the post-hoc analysis model best fits the clean time series derived from the \texttt{RDAE} and \texttt{RAE}. 
This again suggests that \texttt{RAE} and \texttt{RDAE} have the best explainability. 
Next, we consider the concrete example in Fig.~\ref{fig:robustness_example}. 
The key features of the clean time series obtained by \texttt{RDAE}, e.g., the clear periodic pattern, is easier for a human expert to understand what observations should occur at different timestamps. 
In contrast, the clean time series obtained by \texttt{RNNAE} includes many hard-to-understand variations. 
The above observation is justified by a post-hoc analysis--when using \textit{PHE-SSA} with $\gamma=0.1$, we get $N=2$ for \texttt{RDAE} and $N=9$ for \texttt{RNNAE}, suggesting that \texttt{RDAE} has high explainability. %
We also observe that \texttt{RDAE} achieves higher accuracy by having a more explainable clean time series.
See the highlighted parts in Fig.~\ref{fig:robustness_example} that include two outliers. 
\texttt{RDAE} gives two high outlier scores based on its clean time series, making it easy to identify them as outliers, whereas \texttt{RNNAE} gives them small outlier scores. 
This suggests that a more explainable clean time series also contributes improved accuracy. 

\subsubsection{Empirical Convergence Analysis}
\label{ssec:empirical_convergence_analysis}
We study the convergence for (i) \texttt{RAE} with hyperparameters $\lambda$ and (ii) \texttt{RDAE} with with hyperparameter $\lambda$ and $B$.
The convergence analysis is conducted by measuring the difference (i.e., \textit{root mean square error} (\textit{RMSE})) between the original time series $\mathcal{T}$ and the clean time series $\mathcal{T}_{\mathbf{L}}$.
To evaluate the effect of $\lambda$, we vary $\lambda$ among $10^{-4}$, $10^{-3}$, $10^{-2}$, $10^{-1}$, and $1$ while keeping $B=50$. 
To evaluate the effect of $B$, we vary $B$ among $10$, $20$, $50$, $100$, $200$, and $400$ while keeping $\lambda=10^{-1}$. 
We show the results of the convergence analysis for two random time series from dataset \textit{S5} only, due to the space limitation. 
The convergences for the other time series exhibit similar trends.
Fig.~\ref{fig:convergence_analysis_rae_rdae} shows the convergence for \texttt{RAE} and \texttt{RDAE}.
Both \texttt{RAE} and \texttt{RDAE} converge quickly with small $\lambda$ during the first 10 epochs.
Moreover, the convergences are insensitive to $B$ but are sensitive to $\lambda$, and they converge in all cases.

\subsubsection{Runtimes}
\label{ssec:runtime}
We first evaluate the training times of the neural network based methods \texttt{RN}, \texttt{CNNAE}, \texttt{RNNAE}, \texttt{BGAN}, \texttt{DONUT}, \texttt{OMNI}, \texttt{TAE}, \rev{\texttt{RDA}}, \texttt{RAE}, and \texttt{RDAE}. We report findings on dataset \textit{S5} only. The results on the other datasets show similar trends.
Fig.~\ref{fig:runtime_comparision} shows the training \textit{runtime} (\textit{seconds/epoch}). Our methods run extremely fast because they do not perform recursive computations. \rev{\texttt{RDA} and \texttt{RAE} are the fastest methods. \texttt{RDAE} has a very competitive runtime and is only slightly slower than \texttt{CNNAE}.} The other methods run considerably slower due to several reasons: (i) some perform recursive computations (e.g., \texttt{RNNAE} and \texttt{OMNI}); (ii) some are ensemble models that require training of multiple base models (e.g., \texttt{RN}); (iii) some have sampling tricks that take long time (e.g., \texttt{DONUT} and \texttt{OMNI}); (iv) some have complicated objective functions that do not converge easily (e.g., \texttt{BGAN} with the minimax objective function); 
and (v) some have expensive attention score computations (e.g., \texttt{TAE}).
Next, the testing runtimes of the two proposed frameworks and all the other methods are small, i.e., less than 0.1 \textit{seconds}, making them applicable to online outlier detection in streaming settings. 
This evaluation offers evidence that the proposed \texttt{RAE} and \texttt{RDAE} methods can work in time-critical settings while obtaining good accuracy. 

\section{Related Work}
\label{sec:related works}
\noindent
\textbf{Traditional Outlier Detection Methods.}
Traditional methods for outlier detection can be categorized as being distance-based or density-based~\cite{DBLP:books/sp/Aggarwal13}. 
Time series outlier detection is challenging because it is difficult to define a distance or a density metric over the temporal domain. 
Keogh \textit{et al}.~\cite{DBLP:conf/icdm/KeoghLF05} define grammar rules using symbolic representations. 
Matrix Profile~\cite{DBLP:conf/icdm/YehZUBDDSMK16,DBLP:journals/datamine/LinardiZPK20} defines a pair-wise distance between all observations to identify outliers. 
Boniol and Palpanas~\cite{DBLP:journals/pvldb/BoniolP20} propose \texttt{Series2Graph}, which embeds time series into graphs and detects outlier in a graph representation. 

\noindent\textbf{AE-based Methods.}
\texttt{AE}s with 1D \texttt{CNN}s and \texttt{RNN}s have been used for temporal outlier detection~\cite{davidpvldb,DBLP:conf/mdm/Kieu0J18}.
\texttt{AE}s are combined with adversarial training and variational inference for time series anomaly detection~\cite{chen2021daemon,tungicde2022}. 
For spatio-temporal data such as videos, Zhao \textit{et al}.~\cite{DBLP:conf/mm/ZhaoDSLLH17} propose a 3D \texttt{CNN}-based \texttt{AE} to detect outliers in videos. 
\texttt{AE}s are also applied in denoising~\cite{DBLP:conf/icml/VincentLBM08,DBLP:conf/nips/XieXC12}. 
\rev{Compared with an existing robust approach~\cite{DBLP:conf/kdd/ZhouP17}, the proposed frameworks can handle time series, take into account temporal information by using a lagged matrix thus supporting multi-view representations, and support explainability. }
In addition, unlike existing studies that use a single \texttt{AE}, our \texttt{RDAE} framework combines two \texttt{AE}s that help each other identify outliers using two representations of time series data, which improves accuracy. 
\texttt{RDAE} is also different from \texttt{AE} ensembles, where \texttt{RDAE} is a single model that consists of two \texttt{AE}s but an \texttt{AE} ensemble includes many basic models where each basic model is an \texttt{AE}~\cite{DBLP:conf/sdm/ChenSAT17,DBLP:conf/ijcai/KieuYGJ19}. 

\noindent\textbf{RPCA-based Methods.} 
\texttt{RPCA} has been employed to identify outliers~\cite{DBLP:conf/nips/XuCS10}. 
Zhao \textit{et al}.~\cite{DBLP:conf/icml/ZhaoMXZZ14} propose an \texttt{RPCA}-based method to model outliers that follow mixture distributions by combining \texttt{RPCA} with variational inference. 
Lu \textit{et al}.~\cite{DBLP:conf/cvpr/LuFCLLY16} extend \texttt{RPCA} to enable it to work with 3D tensors. 
Fan and Chow~\cite{DBLP:journals/tnn/FanC20} propose Robust Kernel Principal Component Analysis (\texttt{RKPCA}) that combines \texttt{RPCA} with kernel methods to exploit nonlinear mapping functions. %
However, no existing studies apply \texttt{RPCA} to time series. 
Our proposed frameworks are built on \texttt{RPCA}, and the frameworks work for time series. 

\noindent\textbf{Explainable Machine Learning.} 
Explainability is categorized as either intrinsic
or post-hoc~\cite{DBLP:journals/cacm/DuLH20}. 
Most existing explainable methods are proposed for supervised problems~\cite{DBLP:conf/icml/KohL17}. 
Among the intrinsic methods, He \textit{et al}.~\cite{DBLP:conf/aaai/HeLSB19} propose an explainable method for climate prediction. Zhang \textit{et al}.~\cite{DBLP:conf/cvpr/ZhangWZ18a} propose an explainable \texttt{CNN} with filter-object part correspondence. For post-hoc methods, Ribeiro \textit{et al}.~\cite{DBLP:conf/kdd/Ribeiro0G16} propose \texttt{LIME} to explain classifier predictions. Koh and Liang~\cite{DBLP:conf/icml/KohL17} propose a method to trace the model's prediction and back to its training data. 
\rev{Only few explainable methods exist for outlier detection, 
and they often rely on human experts annotation~\cite{DBLP:conf/icdm/YehZUBDDSMK16}, rendering them supervised proposals. These explainable methods focus on indicating the root causes of outliers. Zhang \textit{et al}.~\cite{DBLP:conf/edbt/ZhangDM17} propose a framework to explain outlier events in time series based on annotated events.
Recently, Rad \textit{et al}.~\cite{DBLP:conf/debs/RadSJD21} propose an unsupervised root cause analysis method to explain outliers by indicating the most anomalous dimension in high-dimensional time series. 
In contrast, we quantify the explainability of autoencoder based outlier detection methods rather than explaining individual outlier observations and what causes an outlier observation.
Our explainability analysis method is used to evaluate which autoencoder based outlier detection methods are more explainable, rather than explaining individual outliers.}

\section{Conclusion}
\label{sec:conclusion}
We propose two explainable and robust autoencoder frameworks for unsupervised time series outlier detection. The frameworks represent the first attempt to improve two key aspects of existing neural net based autoencoders: low explainability and high vulnerability to outliers. The frameworks decompose a time series into a clean and an outlier time series, which provides theoretical underpinnings and makes them robust to outliers. We provide a post-hoc explainability analysis method to quantify model explainability. Experimental studies show that the frameworks are effective and outperform strong baselines and state-of-the-art methods. 
In future research, it is of interest to study outlier detection under slightly relaxed settings that are other than fully unsupervised settings, e.g., weakly supervised settings~\cite{SeanIcde2022}. It is also of interest to study different means to further improve outlier detection accuracy, e.g., using ensemble learning~\cite{DBLP:conf/ijcai/KieuYGJ19} and curriculum learning~\cite{DBLP:conf/ijcai/YangGHT021}, and considering time series of location related information. %

\section*{Acknowledgments}
This work was supported in part by Independent Research Fund Denmark under agreements 8022-00246B and 8048-00038B, the VILLUM FONDEN under agreements 34328 and 40567, Huawei Cloud Database Innovation Lab, and the Innovation Fund Denmark center, DIREC.

\bibliographystyle{IEEEtran}
\bibliography{RDA}
\end{document}